\documentclass[11pt]{article}

\usepackage[preprint]{acl}

\usepackage{times}
\usepackage{latexsym}

\usepackage[T1]{fontenc}

\usepackage[utf8]{inputenc}

\usepackage{microtype}

\usepackage{inconsolata}

\usepackage{graphicx}

\usepackage{booktabs}
\usepackage{amsmath}
\usepackage{amssymb}
\usepackage{bbding}
\usepackage{url}
\usepackage{hyperref}
\usepackage[most]{tcolorbox}

\newtcolorbox[auto counter]{findingbox}[1][]{
  enhanced,
  boxrule=0.5pt,
  colframe=teal!60!black,   
  colback=teal!5,           
  arc=1mm,                  
  left=3mm, right=3mm, top=1mm, bottom=1mm,
  before skip=6pt, after skip=6pt,
  before upper=\textbf{\textcolor{teal!70!black}{Finding~\thetcbcounter.\ }}%
  #1                        
}

\newtcblisting{promptbox}[1][]{
  enhanced, 
  breakable,
  title=Prompt,
  colback=gray!6, 
  colframe=gray!60!black,
  colbacktitle=gray!12,
  coltitle=black,
  fonttitle=\bfseries\ttfamily,
  boxrule=0.6pt,
  rounded corners,
  arc=1mm,
  listing engine=listings, 
  listing only,            
  left=2pt,right=2pt,top=4pt,bottom=4pt,
  before upper=\ttfamily\scriptsize, 
  #1
}

\title{Ladder Up, Memory Down: Low-Cost Fine-Tuning 
With Side Nets}

\author{
    Estelle Zheng\textsuperscript{\rm 1, \rm 2},
    Nathan Cerisara\textsuperscript{\rm 1}, \\
    Sébastien Warichet\textsuperscript{\rm 2},
    Emmanuel Helbert\textsuperscript{\rm 2},
    Christophe Cerisara\textsuperscript{\rm 1} \\
    \\
    \textsuperscript{1}Université de Lorraine, CNRS, LORIA, France \\
    \textsuperscript{2}Alcatel-Lucent Enterprise, France
    \\
}

\begin{document}
\maketitle
\begin{abstract}
Fine-tuning large language models (LLMs) is often limited by the memory available on commodity GPUs.
Parameter-efficient fine-tuning (PEFT) methods such as QLoRA reduce the number of trainable parameters, yet still incur high memory usage induced by the backward pass in the full model.
We revisit Ladder Side Tuning (LST), a rarely explored PEFT technique that adds a lightweight side network, and show that it matches QLoRA's compute scaling slope while cutting peak memory by 50\%.  
Across different downstream benchmarks spanning natural language understanding, mathematical and LLM-critic tasks, LST has competitive performance with QLoRA's accuracy on average while being much more memory-efficient. 
This efficiency enables fine-tuning of 7B-parameter models on a single 12 GB consumer GPU with 2k-token contexts, requiring no gradient checkpointing\textemdash conditions under which QLoRA exhausts memory.
Beyond memory efficiency, we also establish scaling laws showing that LST scales similarly to QLoRA.
We exploit Ladder's architectural flexibility by introducing xLadder, a depth-extended variant that increases effective depth via cross-connections and shortens chain-of-thought (CoT) at fixed parameter count.
Ladder is strong when memory is the bottleneck; xLadder builds on this by enabling deeper reasoning without additional memory overhead.

\end{abstract}

\newcounter{takeco}
\setcounter{takeco}{1} 

\section{Introduction}

\begin{figure}[t]
    \centering
    \includegraphics[width=\columnwidth]{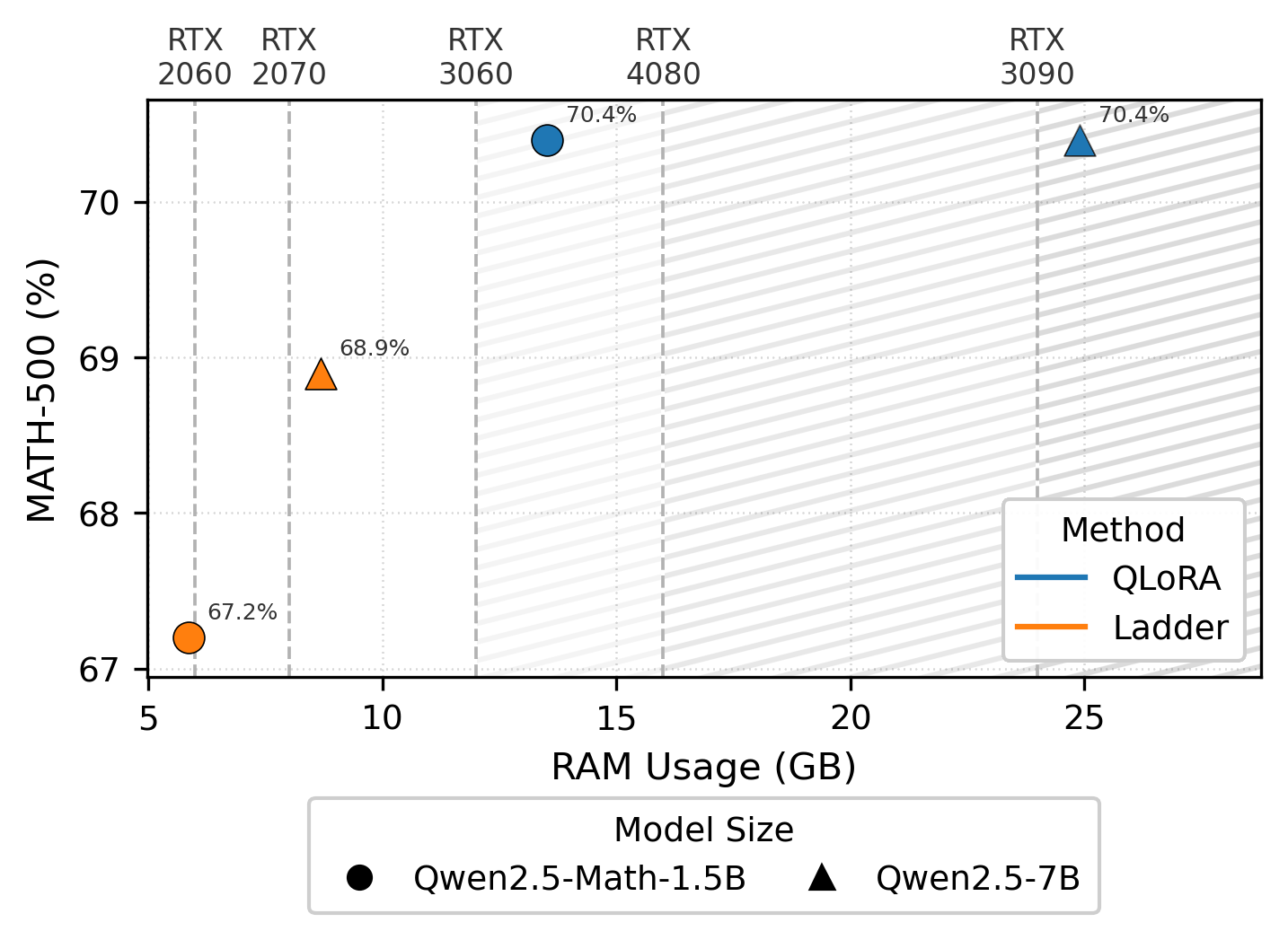}
	\caption{Memory-accuracy frontier on MATH-500. All runs use a batch size of 1 and a 2k-token context window. RAM usage is computed with no gradient checkpointing, as detailed in Appendix~\ref{sec:appendix-vram}. Dashed vertical lines mark the memory budgets of popular consumer GPUs.}
    \label{fig:acc_ram}
\end{figure}


Fine-tuning large language models (LLMs) is crucial for adapting general models
to specific tasks, but it is increasingly limited by GPU memory. LLMs with more than 7B
parameters \cite{grattafiori2024llama3herdmodels,
qwen2025qwen25technicalreport, abdin2024phi4technicalreport} strain commodity
hardware, forcing smaller batches and shorter contexts.
Parameter-efficient fine-tuning (PEFT) reduces the number of trainable
parameters with techniques such as adapters
\cite{li-liang-2021-prefix,lester-etal-2021-power,liu2022fewshot,houlsby2019parameter},
LoRA \cite{hu2022lora} and QLoRA \cite{dettmers2023qlora}, but they still
require backpropagating through the full backbone LLM. As a result, peak memory remains
dominated by activation storage rather than weights, making 
fine-tuning dependent on gradient checkpointing or multiple GPUs. 

\begin{figure}[htbp]
    \centering
    \includegraphics[width=\columnwidth]{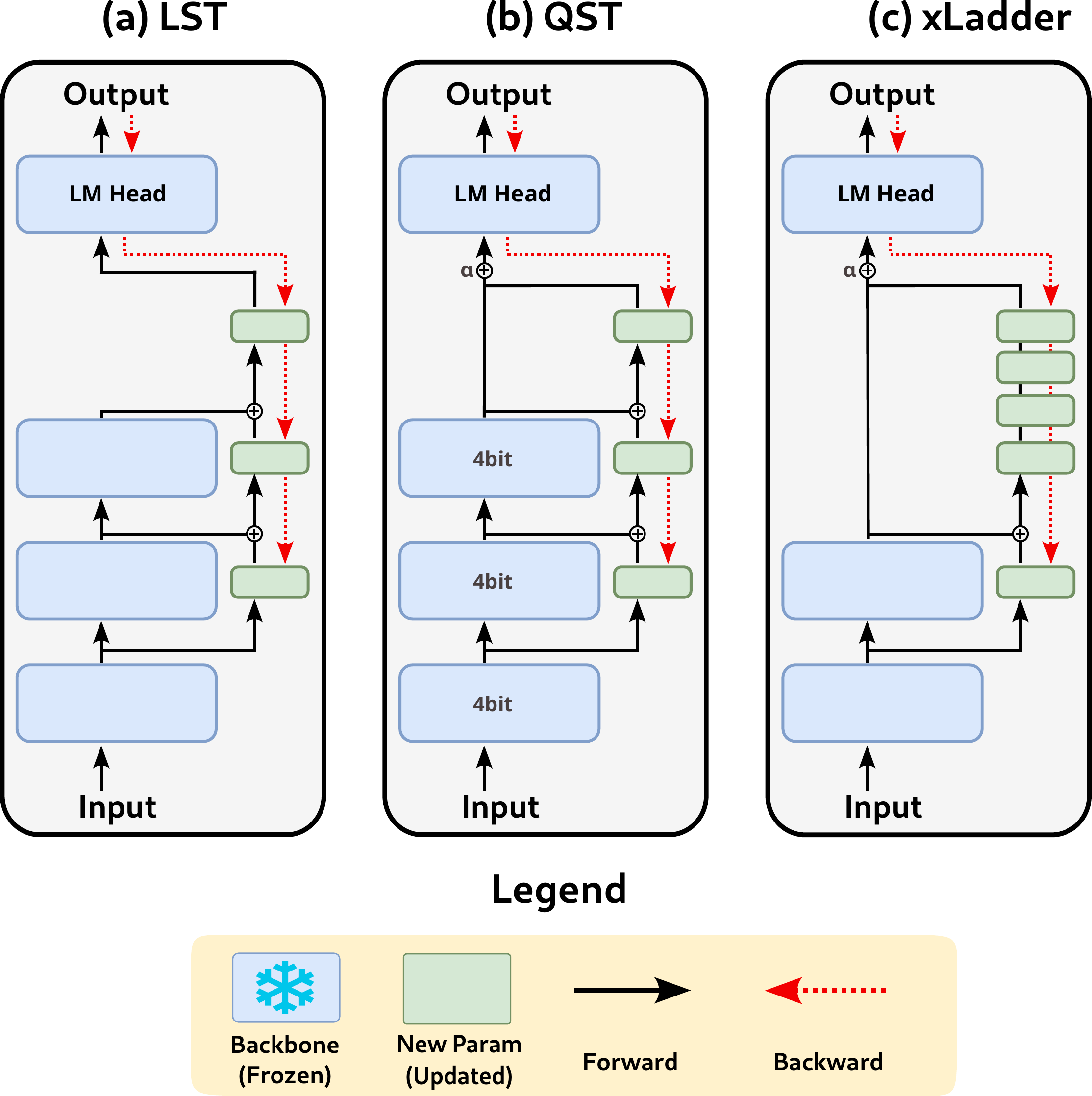}
    \caption{Architecture comparison of Ladder-based methods.}
    \label{fig:sota_ladder}
\end{figure}

Ladder-based side tuning takes a different path: it removes backward passes through the backbone, cutting peak memory by roughly half. As shown in Figure~\ref{fig:sota_ladder}, Ladder Side Tuning (LST) \cite{sung2022lst} connects each Transformer block with lightweight projections into a shallow side network. Gradients flow only through the side path, keeping the backbone in inference mode and discarding intermediate activations on the fly, which typically halves memory (Figure~\ref{fig:acc_ram}). Quantized Side Tuning (QST) \cite{zhang-etal-2024-quantized} adds 4-bit quantization to LST.

Despite their computational promise, ladder methods remain under-explored in the literature. Prior studies reported empirical gains on isolated tasks but left two fundamental gaps: (i) no theoretical analysis of the memory savings achieved under generation tasks, and (ii) no scaling-law analysis to understand how ladder methods behave as data, model size, or compute resources grow. 

In this work, we bring two main novel insights about the Ladder family of methods:
\begin{enumerate}
	\item \textbf{Scaling \& efficiency.} Across compute, memory and downstream accuracy, Ladder scales reliably well; its main benefit is memory savings.
	\item \textbf{Architectural flexibility.} Ladder architecture can be extended into a general design framework that supports ensembling, cascading, and depth extension, enabling targeted variants for different use cases.
\end{enumerate}
In the following, unless specified otherwise, Ladder denotes the original LST fully connected architecture, as the QST variant only adds quantization, resulting in slightly worse performance \cite{zhang-etal-2024-quantized}.

\section{Related Work}
\textbf{Parameter-Efficient Fine-Tuning (PEFT).}
Full fine-tuning of LLMs is often impractical due to memory and compute costs. PEFT mitigates this by training small modules on top of a frozen backbone, such as prompt/prefix tuning \cite{lester-etal-2021-power, li-liang-2021-prefix}, IA$^{3}$ \cite{liu2022fewshot}, and LoRA \cite{hu2022lora}. QLoRA further combines LoRA with 4-bit weight quantization to cut memory for weights and optimizer states \cite{dettmers2023qlora}.


\textbf{Ladder and Side-Style Tuning.}
An orthogonal line of work replaces backpropagation through the backbone with a lightweight side network. Ladder Side Tuning (LST) \cite{sung2022lst} attaches linear projections to each Transformer block and trains only the side path, thereby discarding intermediate activations. 
Subsequent work has extended and refined this paradigm. Quantized Side Tuning (QST) \cite{zhang-etal-2024-quantized} adds 4-bit quantization to scale the approach to larger models. Calibration Side Tuning (CST) \cite{chen2024cst} introduces tiny decoder-block calibrators specifically for vision-language transfer tasks. Low-rank Attention Side Tuning (LAST) \cite{tang2024lowrankattentionsidetuningparameterefficient} replaces the side MLP with a low-rank attention projector for encoder-decoder tasks. We step back from prior work and study fundamental properties of these architectures.

\noindent\textbf{Memory-Efficient Fine-Tuning.}
Activation memory represents a major bottleneck when fine-tuning large models, particularly with long contexts. It can be reduced by recomputing, with gradient checkpointing \cite{chen2016trainingdeepnetssublinear} or by using reversible layers \cite{gomez2017reversible,kitaev2020reformer}.  
ZeRO-offload \cite{ren2021zero} offloads optimizer states to CPU memory.
Network compression methods such as pruning \cite{frankle2020linear}, and
distillation \cite{sanh2020distilbert,hinton2015distilling} shrink the
inference model, but still require storing activations during training. Some
approaches bypass backpropagation by using only forward passes
\cite{malladi2023mezo}
, but at the cost of noisier gradients or training steps.
These gradients approximation approaches are actually complementary to
Ladder methods, which use a highly reduced set of gradients, either exact or approximate.

\noindent\textbf{Fine-Tuning for Reasoning. }
Recent advances in math and multi-step reasoning have relied on reinforcement
learning (RL) \cite{chu2025sft,lightman2024lets, yuan2025free} or self-play
\cite{chen2024selfplay} to improve LLMs math reasoning abilities. The use of
Supervised Fine-Tuning (SFT) with long distill traces from larger models enables
training reasoning LLMs at lower cost than with RL
\cite{muennighoff2025s1simpletesttimescaling,
ye2025limoreasoning,guha2025openthoughtsdatarecipesreasoning,deepseekai2025deepseekr1incentivizingreasoningcapability}.
With larger and larger contexts to train on, Ladder methods can prove useful to
reduce the memory footprint of such SFT reasoning LLMs.

\noindent\textbf{Scaling Laws on Supervised Fine-Tuning (SFT).}
While power-law scaling relationships are well-established for pretraining \cite{kaplan2020scalinglawsneurallanguage, hoffmann2022an} extending these analyses to fine-tuning presents unique challenges. Several recent works have begun establishing fine-tuning and PEFT scaling laws \cite{gadre2024languagemodelsscalereliably, bhagia2024establishingtaskscalinglaws, wyatte2024scaling, yizthak2024languagemodelsscalereliably,zhang2024when, qi2025evolmsearchlostlanguage}, but without considering ladder architectures:
given their major differences with other PEFT adapters, it is far from obvious whether scaling laws are the same. We give insights to answer this open question next. 

\section{Architectural flexibility}
\label{sec:methodology}
Ladder Side Tuning (LST) can be understood not only as a low-memory, parameter-efficient fine-tuning strategy, but also as a general interface for coupling a large, frozen $L$-layer transformer with a smaller, trainable $l$-layer transformer.  
Let $\mathcal{C}$ denote the set of cross-layer connections between the two stacks.  Different choices of $\mathcal{C}$ recover familiar model-combination paradigms:
\begin{itemize}
    \item \textbf{Full ladder}. When $L{=}l$ and $\mathcal{C}_{\text{full}} = \{i \rightarrow i\}_{1\le i\le L}$, we obtain the original fully connected ladder of \citet{sung2022lst}.
    \item \textbf{Ensembling}. Setting $\mathcal{C}_{\text{ens}} = \{1\rightarrow 1,\; L\rightarrow l\}$ is equivalent to averaging the outputs of two independent transformers.
    \item \textbf{Cascading}. With $\mathcal{C}_{\text{cas}} = \{L\rightarrow 1\}$, the frozen backbone feeds its final representation to the first layer of the small model, yielding a standard cascade.
\end{itemize}

The ladder framework offers significant design flexibility, supporting both sparse and dense configurations, as well as parallel and crossing connections. This makes it a valuable testbed for analyzing the impact of inter-model information flow.

We explore an \textit{extended ladder}, or \textbf{xLadder}, that adds extra reasoning depth while still leveraging the mid-level abstractions from the backbone, see Figure~\ref{fig:sota_ladder}. Specifically, we define
\begin{equation*}
\label{eq:cxladder}
\resizebox{\columnwidth}{!}{$
    \mathcal{C}_{\text{xladder}} 
    = \{(L-\delta) \rightarrow 1,\; (L-\delta+1) \rightarrow 2,\; \dots,\; L \rightarrow \delta \}
$}
\end{equation*}

where $\delta = l/2$ by default. This yields an \smash{$(L+\delta)$}-layered computation graph: the frozen backbone contributes its last $\delta$ layers in parallel with the first $\delta$ layers of the lightweight ladder, effectively deepening the model without additional backbone training cost. 

Middle layers of LLMs are known to encode more abstract, syntax-agnostic
concepts, which have been shown to enhance reasoning capabilities
\cite{jawahar-etal-2019-bert,voita-etal-2019-bottom}. The first and last layers
focus on surface-level lexical patterns. By integrating those mid-level
abstractions, xLadder may increase the number of latent reasoning steps, hence
reducing the need for long, complex chain-of-thoughts (CoT).
Such exchange of layers for tokens is actually promoted in recent reasoning LLMs, such as GLM-4.5~\footnote{see \url{https://z.ai/blog/glm-4.5}}.
Under the assumption that the extra side depth $\delta$ is kept fixed (or grows sublinearly) w.r.t. model scale, xLadder inherits the same memory and compute scaling as Ladder, up to constant factors.

\section{Scaling Laws}
\subsection{Experimental Setup}
\textbf{Datasets.}
We train on the EvoLM math reasoning corpus
\cite{qi2025evolmsearchlostlanguage}, a 500,000 instance collection created by augmenting the MATH
\cite{hendrycks2021measuringmathematicalproblemsolving} and GSM8K
\cite{cobbe2021trainingverifierssolvemath} with problems extracted from MetaMathQA
\cite{yu2024metamathbootstrapmathematicalquestions}, OpenMathInstruct2
\cite{toshniwal2025openmathinstruct} and NuminaMath
\cite{numina_math_datasets}. EvoLM is released at multiple predefined subset sizes, ranging from 
100k to 500k samples, enabling the derivation of scaling laws, using the original math problems 
and their solutions. Unlike several recent math reasoning datasets, EvoLM does not contain any
intermediate trace produced by larger teacher models; therefore, no distillation is involved,
and SFT only exploits the original reasoning traces that are provided within the datasets.

For downstream performance evaluation, we additionally fine-tune on two GLUE
\cite{wang-etal-2018-glue} tasks: CoLA (Corpus of Linguistic Acceptability) and
QQP (Quora Question Pairs). Both are binary classification problems, CoLA
assesses grammatical acceptability, whereas QQP tests semantic equivalence of
question-answer pairs. Together, these datasets contribute roughly to 800k
samples, which is sufficient to derive scaling laws.

\noindent\textbf{Models.}
We experiment primarily with models from the Qwen family, including Qwen-2.5-Math~1.5B \cite{yang2024qwen25mathtechnicalreportmathematical}, and Qwen-2.5~7B \cite{qwen2025qwen25technicalreport}. 
For downstream tasks, we evaluate from the Qwen non-math, Llama and OPT series, such as Qwen-2.5 0.5B, 1.5B, 3B \cite{qwen2025qwen25technicalreport}; Llama-3.2 1B, 3B \cite{grattafiori2024llama3herdmodels} and OPT-1.3B, 2.7B, 6.7B \cite{zhang2022optopenpretrainedtransformer}.
The Qwen models are chosen for their strong performance on reasoning tasks, while the Llama and OPT models provide a broader range of parameter sizes and architectures for comparison.

\noindent\textbf{Implementation.}
The model is trained on subsets of the EvoLM dataset \cite{qi2025evolmsearchlostlanguage} that range from 100k to 500k samples to plot scaling laws: see Appendix~\ref{sec:isoflops} for details. The best model is chosen based on the validation subset of EvoLM. The prompt template encourages the answer to be in the \texttt{\textbackslash boxed\{\}} format. The exact prompt used can be found in Appendix~\ref{sec:appendix-math-prompt}. The model is trained on a single NVIDIA A100-80GB GPU.
More details on the training hyperparameters, model configurations and hardware can be found in Appendix~\ref{sec:appendix-scaling-hyperparam} and ~\ref{sec:appendix-scaling-ladder-config}.

For the classification tasks, 
the best model is selected based on early stopping on the validation loss.
We run each experiment five times for CoLA and twice for QQP, and we report the
average score. More details on hyperparameters can be found in Appendix~\ref{sec:appendix-downstream-hyperparam}.

\subsection{Test Loss Scaling Laws}
\label{sec:scaling_loss}
\citet{zhang2024when} derive, among others, the following scaling law for LoRA fine-tuning:
\begin{equation}
    \label{eq:scaling_law_paper}
	\hat{\mathcal{L}}(D_f) = \frac {A}{D_f^{\lambda}} + E
\end{equation}
This law, parameterized by scalars $A$,~$\lambda$ and~$E$, describes how the fine-tuning loss $\hat{\mathcal{L}}(D_f)$ varies with the dataset size $D_f$.
However, fair comparison of PEFT methods should be done at same data and compute.
Let $\zeta_m$ be the number of floating-point operations (FLOPs) per token of a given PEFT method $m$ 
with $n$ trainable parameters applied on a backbone LLM of size $N$.
Since QLoRA backpropagates through the backbone while the Ladder only backpropagates
through the side network, the standard FLOPs estimates are~\cite{kaplan2020scalinglawsneurallanguage}:
\[
\begin{aligned}
\zeta_{\text{QLoRA}} &= 6(N + n) \\
\zeta_{\text{Ladder}} &= 2N + 6n
\end{aligned}
\]
The total training compute
$C = \zeta_m \times D_f$ can be substituted into Eq.~\eqref{eq:scaling_law_paper} to get a compute-based law:
\begin{equation}
    \label{eq:scaling_law}
	\mathcal{L}(C) = \frac{A\zeta_m^{\lambda}}{C^{\lambda}} + E = \frac{B}{C^{\lambda}} + E
\end{equation}
with $B=A\zeta_m^{\lambda}$.
The backbone is frozen ($N$ is constant) and $C$ depends on $D_f$, which varies.
$n$ is estimated as usual from isoFLOPs (see Appendix~\ref{sec:isoflops}).



\begin{figure}[htbp]
    \centering
    \includegraphics[width=\columnwidth]{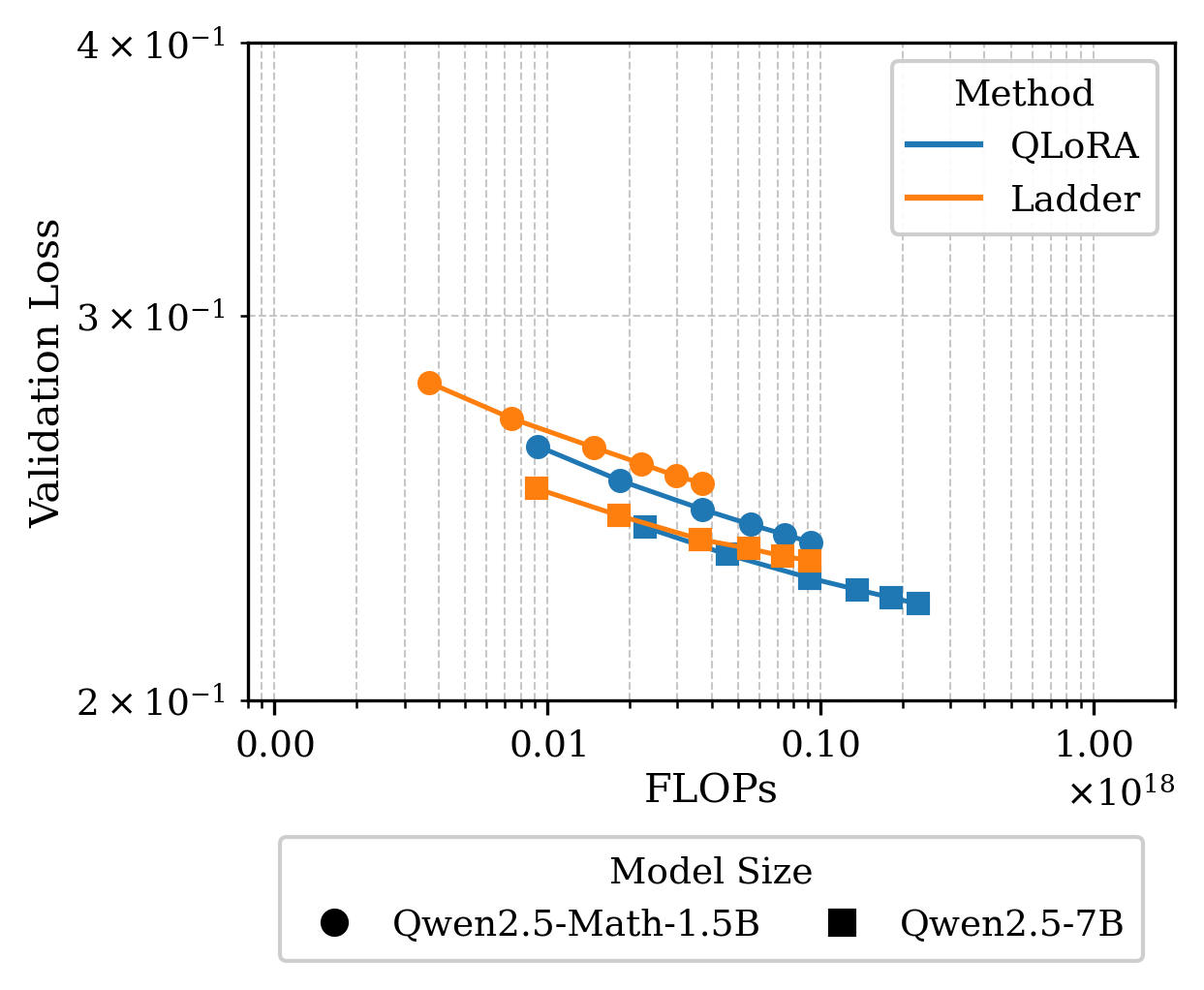}
    \caption{Scaling laws of QLoRA and Ladder methods.}
    \label{fig:scaling_loss}
\end{figure}




Using Eq.~\eqref{eq:scaling_law} and the curves in Figure~\ref{fig:scaling_loss}, we estimate the scaling law coefficients for both Ladder and QLoRA methods by means of nonlinear least squares regression and Huber loss to better validated the method, as reported in Table~\ref{tab:scaling_nls}.

\begin{table}[htbp]
\centering
\resizebox{.85\width}{!}{%
\begin{tabular}{@{}l c c c@{}}
\toprule
\textbf{Model} & $\boldsymbol{\lambda}$ & $\boldsymbol{B}$ & $\boldsymbol{E}$ \\ \midrule
\multicolumn{4}{c}{\textbf{\textit{Curve Fit}}} \\ \midrule
QLoRA-Qwen-1.5B & 0.26 \scriptsize{$[0.21,0.31]$} & 869 & 0.21 \scriptsize{$\pm 0.01$} \\
QLoRA-Qwen-7B   & 0.27 \scriptsize{$[0.24,0.31]$} & 695 & 0.20 \scriptsize{$\pm 0.00$} \\ \midrule
Ladder-Qwen-1.5B & 0.22 \scriptsize{$[0.12,0.31]$} & 189 & 0.21 \scriptsize{$\pm 0.01$} \\
Ladder-Qwen-7B   & 0.26 \scriptsize{$[0.10,0.48]$} & 534 & 0.21 \scriptsize{$\pm 0.02$} \\ \midrule
\multicolumn{4}{c}{\textbf{\textit{Huber Loss Fit}}} \\ \midrule
QLoRA-Qwen-1.5B & 0.26 \scriptsize{$[0.18,0.29]$} & 701 & 0.21 \scriptsize{$\pm 0.02$} \\
QLoRA-Qwen-7B   & 0.26 \scriptsize{$[0.15,0.28]$} & 698 & 0.20 \scriptsize{$\pm 0.02$} \\ \midrule
Ladder-Qwen-1.5B & 0.22 \scriptsize{$[0.07,0.28]$} & 189 & 0.21 \scriptsize{$\pm 0.03$} \\
Ladder-Qwen-7B   & 0.26 \scriptsize{$[0.21,0.52]$} & 546 & 0.21 \scriptsize{$\pm 0.02$} \\\bottomrule
\end{tabular}%
}
\caption{Test loss scaling on different Qwen models for QLoRA and Ladder. 95\% Bootstrap CI is reported for $\lambda$ and std is reported for $E$.}
\label{tab:scaling_nls}
\end{table}

Figure~\ref{fig:scaling_loss} and Table~\ref{tab:scaling_nls} show that the scaling slopes $\lambda$ for both methods are very similar, indicating closely matched scaling properties. While the compute efficiency $B$ differs, this mainly reflects vertical shifts in the loss-compute curves without altering the scaling behavior itself. The asymptotic floor $E$ is also comparable across methods and sizes, suggesting a data-dominated regime for a fixed backbone.

\begin{findingbox}
Despite the lack of gradients in the backbone LLM, Ladder scales as well as QLoRA.
\end{findingbox}



\subsection{Downstream Performance Scaling Law}
\label{sec:scaling_acc}
Prior works~\cite{qi2025evolmsearchlostlanguage, gadre2024languagemodelsscalereliably, bhagia2024establishingtaskscalinglaws, wyatte2024scaling} have shown that LLMs' fine-tuning performance is preferably measured with accuracy rather than loss when studying their scaling properties.
We derive from the error power law proposed by~\citet{yizthak2024languagemodelsscalereliably} the following scaling relation on downstream tasks:
\begin{equation}
    \label{eq:scaling_acc}
    Err(C) = \kappa C^{-\beta} + E_{\infty} \text{, with } Acc = 1 - Err(C)
\end{equation}
where $Err(C)$ is the error, $C$ is the compute budget, and $\kappa$, $\beta$, and $E_{\infty}$ are parameters to fit. 

\begin{figure}[htbp]
    \centering
    \includegraphics[width=\columnwidth]{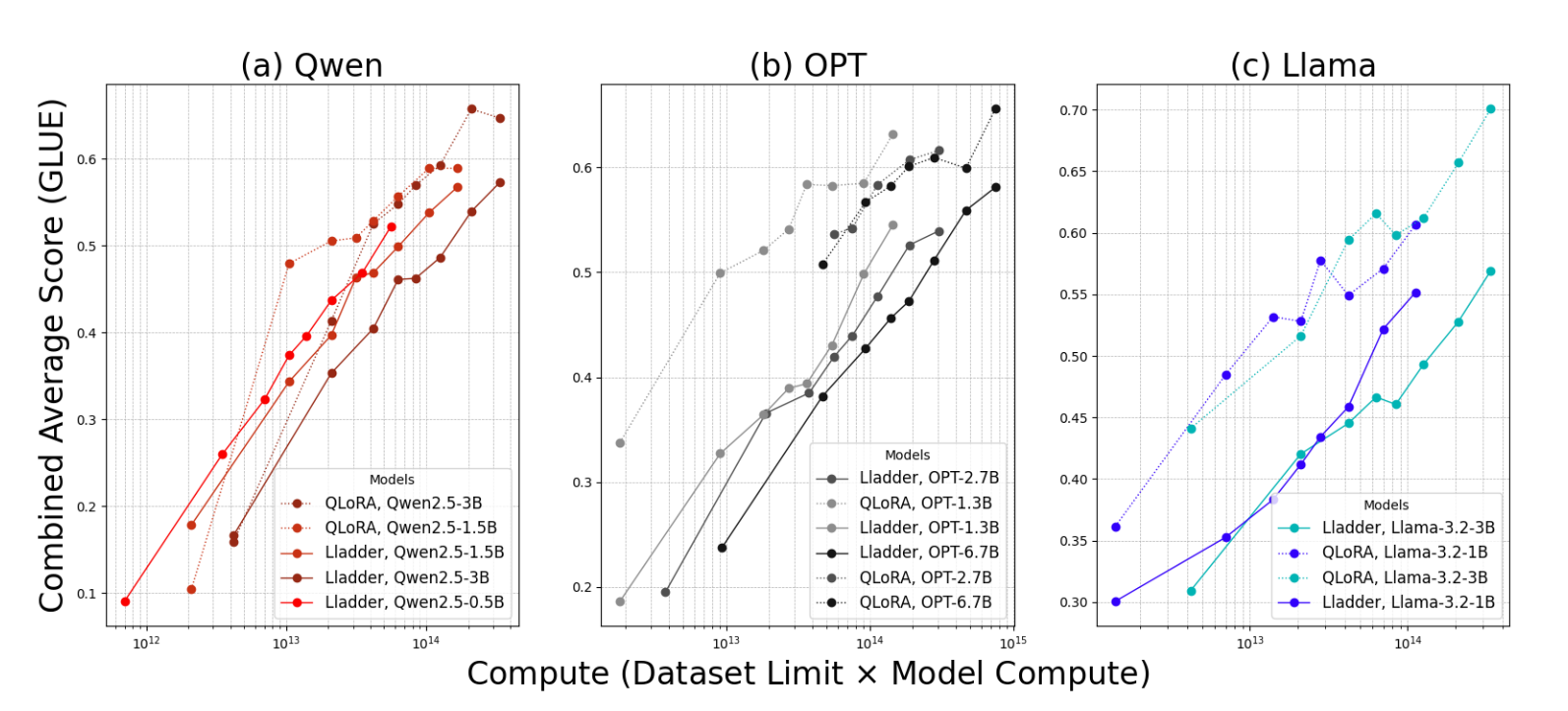}
    \caption{Scaling relation between the Ladder compute and the downstream tasks accuracy for 3 LLM series: (a) Qwen models, (b) OPT models, (c) LLama models.}
    \label{fig:scaleacc}
\end{figure}

We fit Eq.~\eqref{eq:scaling_acc}, estimating the scaling exponent $\beta$, the asymptotic error $E_{\infty}$, and the prefactor $\kappa$. Across methods and sizes, the inferred ceilings are near the task maximum within uncertainty, so we focus on $\beta$; $\kappa$ primarily induces vertical shifts, compute efficiency, without altering exponent-driven scaling behavior.

Figure~\ref{fig:scaleacc} suggests similar slopes in the mid-compute regime. Full estimates appear in Appendix~\ref{sec:appendix-scaling-fit}.

\begin{findingbox}
    Ladder offers comparable accuracy-per-compute to QLoRA, but with a more stable scaling slope.
    Therefore, the Ladder neither improve nor degrade the performance-training speed ratio when compared to QLoRA.
\end{findingbox}



\subsection{Memory Scaling Relations}
\label{sec:memory-scaling}
Fine-tuning large pretrained models is memory-intensive. The training-time GPU memory (VRAM) can be decomposed into three main terms: M$_1$: model weights, M$_2$: optimizer states, and M$_3$: intermediate activations. PEFT methods aim to cut training cost by shrinking one or more of these terms. 

LoRA \cite{hu2022lora} lowers M$_2$ by updating only a small set of low-rank adapter parameters, while QLoRA \cite{dettmers2023qlora} additionally reduces M$_1$ by quantizing the (frozen) backbone weights. However, in both cases the model's intermediate activations M$_3$ are still materialized in (near) full precision, which dominates the memory footprint. 
In practice, M$_3$ is often the largest term during fine-tuning.

\begin{figure}[htbp]
    \centering
    \includegraphics[width=\columnwidth]{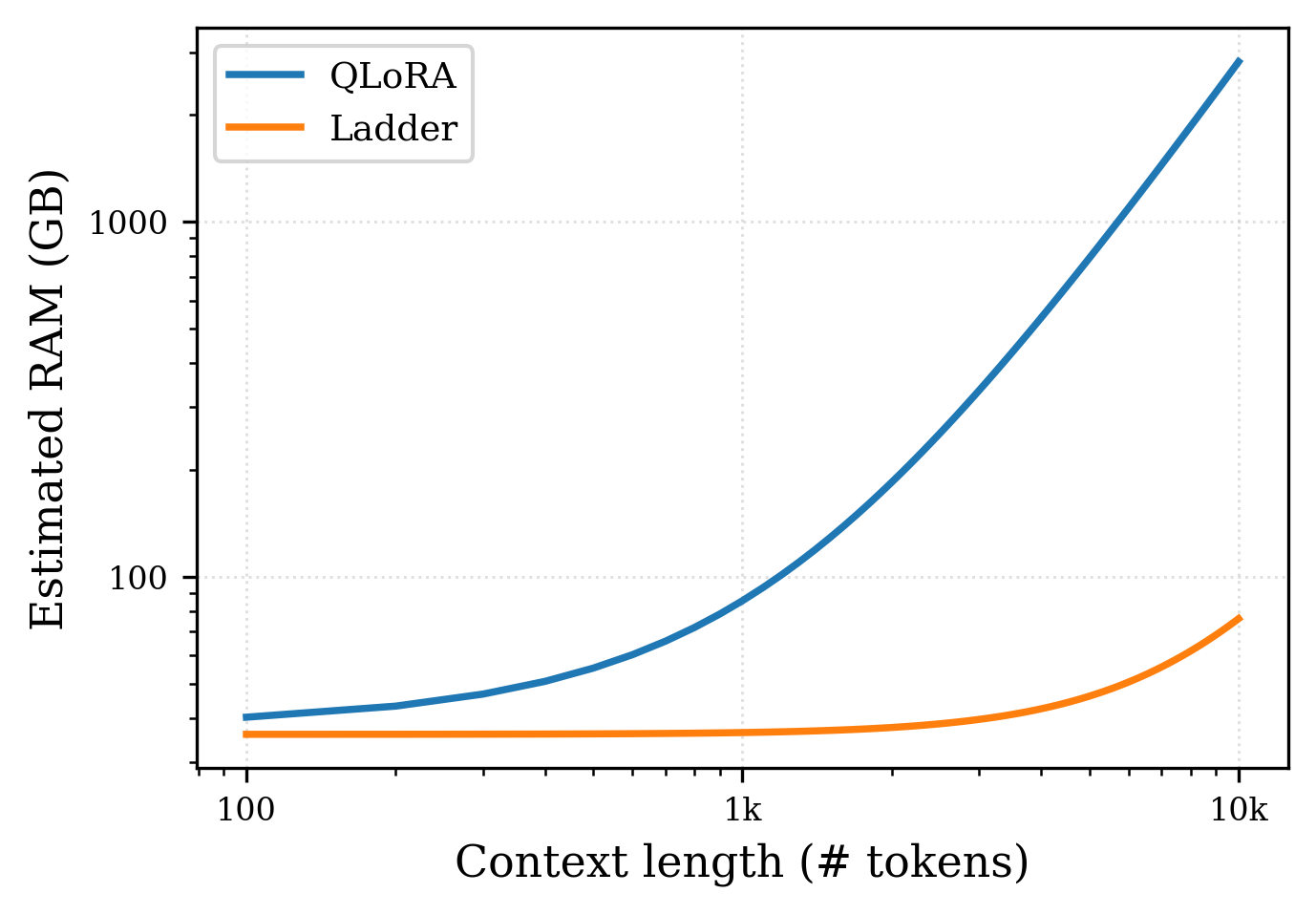}
	\caption{GPU memory required by QLoRA and Ladder during training of a 70B-parameter decoder-only transformer. We assume 8-bit AdamW, vanilla attention, and no gradient checkpointing. See Appendix~\ref{sec:appendix-vram} for the full derivation.}
    \label{fig:vram}
\end{figure}

Ladder approaches such as LST \cite{sung2022lst} and QST \cite{zhang-etal-2024-quantized} use a side lightweight model to reduce the memory footprint of the optimizer state (M$_2$) and intermediate activations (M$_3$). QST further reduces M$_1$ by quantizing the backbone model weights, while LST maintains full precision backbone weights. The resulting memory savings are derived in Appendix~\ref{sec:appendix-vram} and visualized in Figure~\ref{fig:vram}.

\begin{table*}[htbp]
\centering
\resizebox{.85\width}{!}{%
\begin{tabular}{@{}lccccccc@{}}
    \toprule
    \textbf{Model} & \textbf{Pure SFT?} & \textbf{MATH-500} & \textbf{AIME24} & \textbf{AIME25} & \textbf{AMC23} & \textbf{Minerva} & \textbf{\begin{tabular}[c]{@{}c@{}}Olympiad\\ Bench\end{tabular}} \\ \midrule
    \multicolumn{8}{c}{\textit{Based on Qwen2.5-7B}} \\ \midrule
    Qwen Base \scriptsize{\cite{qwen2025qwen25technicalreport}} & \Checkmark & 61.2 \footnotesize{$\pm$ 0.4} & 8.7 \footnotesize{$\pm$ 2.4} & 7.9 \footnotesize{$\pm$ 4.3} & 32.5 \footnotesize{$\pm$ 1.5} & 16.9 \footnotesize{$\pm$ 1.2} & 30.2 \footnotesize{$\pm$ 2.3}  \\
    Qwen Instruct \scriptsize{\cite{qwen2025qwen25technicalreport}} & \XSolidBrush & 75.2 \footnotesize{$\pm$ 0.5} & 8.7 \footnotesize{$\pm 2.4$} & 8.7 \footnotesize{$\pm 4.3$} & 38.5 \footnotesize{$\pm$ 1.4} & 35.8 \footnotesize{$\pm 1.0$} & 38.7 \footnotesize{$\pm 1.0$}  \\
    s1.1-7B ($\ast$) \scriptsize{\cite{muennighoff2025s1simpletesttimescaling}}& \XSolidBrush & 80.8 \footnotesize{$\pm$ 0.6} & 19.0 \footnotesize{$\pm$ 3.2} & 21.0 \footnotesize{$\pm$ 5.5} & 59.5 \footnotesize{$\pm$ 3.7} & 37.5 \footnotesize{$\pm$ 1.1} & 48.2 \footnotesize{$\pm 1.4$} \\
    Full SFT ($\ast$) \scriptsize{\cite{wang2025unleashingreasoningpotentialpretrained}}  & \XSolidBrush & 58.6  & 10.0 & 7.1 & 45.3 & 24.6 & 27.6  \\
    QLoRA & \Checkmark  & {\textbf{70.4}} \footnotesize{$\pm$ 1.4} & {\textbf{8.7}} \footnotesize{$\pm$ 1.8} & 6.0 \footnotesize{$\pm$ 4.3} & 41.5 \footnotesize{$\pm$ 1.4} & {\textbf{33.5}} \footnotesize{$\pm$ 1.3} & 32.0 \footnotesize{$\pm$ 0.8} \\
    Ladder (ours) & \Checkmark & 68.9 \footnotesize{$\pm$ 1.6} & 8.0 \footnotesize{$\pm$ 1.8} & {\textbf{8.0}} \footnotesize{$\pm$ 1.8} & {\textbf{47.5}} \footnotesize{$\pm$ 7.3} & 29.3 \footnotesize{$\pm$ 1.2} & {\textbf{34.3}} \footnotesize{$\pm$ 0.5} \\ \midrule
    \multicolumn{8}{c}{\textit{Based on Qwen2.5-Math-1.5B}} \\ \midrule
    
    Qwen Base \scriptsize{\cite{yang2024qwen25mathtechnicalreportmathematical}} & \Checkmark & 42.0 \footnotesize{$\pm$ 4.7} & 11.3 \footnotesize{$\pm$ 3.6} & 5.7\footnotesize{$\pm$ 2.1} & 37.5 \footnotesize{$\pm$ 3.5} & 12.1 \footnotesize{$\pm$ 1.7} & 23.8 \footnotesize{$\pm$ 0.8}  \\
    Qwen Instruct \scriptsize{\cite{yang2024qwen25mathtechnicalreportmathematical}} & \XSolidBrush & 74.8 \footnotesize{$\pm$ 0.5} & 8.7 \footnotesize{$\pm$ 2.4} & 7.3 \footnotesize{$\pm$ 5.1} & 42.0 \footnotesize{$\pm$ 2.5} & 31.6 \footnotesize{$\pm$ 5.7} & 37.9 \footnotesize{$\pm$ 2.2}  \\
    Full SFT ($\ast$) \scriptsize{\cite{wang2025unleashingreasoningpotentialpretrained}} & \Checkmark & 49.0 & 7.9 & 2.1 & 35.8 & 14.3 & 23.2 \\
     QLoRA & \Checkmark & {\textbf{70.4}} \footnotesize{$\pm$ 0.5} & 6.7 \footnotesize{$\pm 2.4$} & 6.7 \footnotesize{$\pm$ 3.3} & 45.0 \footnotesize{$\pm$ 2.5} & 29.1 \footnotesize{$\pm$ 0.6} & {\textbf{33.7}} \footnotesize{$\pm$ 1.6} \\
    Ladder (ours) & \Checkmark & 67.2 \footnotesize{$\pm 1.4$} & {\textbf{8.0}} \footnotesize{$\pm$ 5.1} & {\textbf{8.0}} \footnotesize{$\pm$ 3.0} & {\textbf{46.0}} \footnotesize{$\pm$ 4.9} & {\textbf{29.4}} \footnotesize{$\pm$ 1.8} & 32.7 \footnotesize{$\pm$ 1.8} \\ \bottomrule
\end{tabular}%
}
\caption{Experiments results on different math reasoning benchmark. We report Pass@1 accuracy (mean $\pm$ std) of all methods across 5 random seeds. We report in ($\ast$) the results from their original papers with their corresponding std. We note that some RL-based models are used as baselines, while ours is solely based on SFT.}
\label{tab:math_results}
\end{table*}

To ensure a fair comparison, we evaluate QLoRA and Ladder under the same training configuration (identical batch size and sequence length). 
Figure~\ref{fig:acc_ram} shows that, at comparable model sizes, 
QLoRA slightly outperforms Ladder on MATH-500, but at the cost of more compute (see Section~\ref{sec:scaling_loss})
and substantially higher VRAM. In practice, fine-tuning a 7B model with Ladder becomes possible
on low-end consumer GPUs, while it may be impossible with QLoRA on the same GPU. 
Although memory can be traded for compute with gradient checkpointing (see
Appendix~\ref{sec:appendix-vram} for plots that include gradient checkpointing),
the memory gap between QLoRA and Ladder remains significant regardless of the memory optimization method used.
To keep comparisons fair, we restrict our analysis to vanilla attention.
There are many memory-saving techniques (Sliding Window Attention, Grouped Query Attention (GQA) \cite{ainslie-etal-2023-gqa}, Flash Attention1-3 \cite{dao2022flashattentionfast}, PagedAttention \cite{kwon2023efficient}, RadixAttention \cite{zheng2024sglang}, FlexAttention \cite{dong2024flexattentionprogrammingmodel},
caching and NVMe offloading~\footnote{See, e.g., \url{https://github.com/Mega4alik/ollm}}),
but they differ in assumptions and costs, so a comprehensive comparison is beyond this section.

\begin{findingbox}
    Ladder methods offer significantly better memory efficiency than QLoRA, enabling fine-tuning large models on low-end consumer-grade GPUs.
\end{findingbox}



\section{Evaluation on Downstream Tasks}

\subsection{Math Reasoning Tasks}

\textbf{Setup.}
We train a Ladder adapter on the 1,000 training samples of the s1K-1.1 dataset
\cite{muennighoff2025s1simpletesttimescaling}, which is a curated version of
59k math reasoning examples from multiple sources and using the trace of larger
reasoning models as answer. The context length on this dataset is limited to 2,122 tokens corresponding to the longest example in the training set.
Hyperparameters are chosen based on the validation
loss and accuracy on a subset of the EvoLM dataset
\cite{qi2025evolmsearchlostlanguage}. The prompt template used for training
and evaluation follows the answer in a \texttt{\textbackslash boxed\{\}}
format. 

We evaluate on the MATH-500 dataset \cite{hendrycks2021measuringmathematicalproblemsolving}, which is a benchmark comprising competition math problems of varying difficulty (we evaluate on the same 500 samples selected by OpenAI in prior work \cite{lightman2024lets}), and on the AIME'24 and AIME'25 datasets \cite{aime}, each containing 30 high-school-level math problems. We also evaluate on the AMC'23 dataset \cite{amc}, with 40 examples from math competitions, and on the Minerva Math dataset \cite{lewkowycz2022solving}, containing 272 undergraduate-level problems; and on the OlympiadBench dataset \cite{he-etal-2024-olympiadbench}, comprising 674 examples.

Following \citet{hochlehnert2025soberlookprogresslanguage} on evaluation metrics, we report the Pass@1 accuracy, which is the percentage of problems solved correctly on the first attempt.
All datasets are evaluated over 5 random seeds to ensure robustness, and the results are averaged. 
We evaluate with greedy decoding with \texttt{max\_new\_tokens} of 2,048 tokens using Math-Verify~\cite{mathverify} consistent with prior work indicating this budget is sufficient on these datasets~\cite{yang-etal-2025-markov}.


We use Qwen series models such as Qwen~2.5~Math~1.5B \cite{yang2024qwen25mathtechnicalreportmathematical}, and Qwen~2.5~7B \cite{qwen2025qwen25technicalreport} as backbone models. More details on the training hyperparameters can be found in Appendix~\ref{sec:appendix-implementation-math}.

\noindent\textbf{Results.}
Table~\ref{tab:math_results} contrasts our Ladder models with QLoRA and other published baselines on several math reasoning benchmarks. 
Scores for s1.1-7B are taken directly from the Sober leaderboard \cite{hochlehnert2025soberlookprogresslanguage}, while the Full SFT results are from \citet{wang2025unleashingreasoningpotentialpretrained} which does not report standard deviations. 
Both experimental setups slightly differ from ours, particularly in their generation token length of 32,768 tokens, which is significantly longer than our 2,048 tokens. 
Despite these differences, our Ladder achieves competitive performance compared to QLoRA.

Prior works suggest that RL-based methods can yield larger gains than SFT-based methods on reasoning tasks \cite{chu2025sft,luo2025wizardmath,trung-etal-2024-reft}. Combining the Ladder with RL might be a promising solution to democratize reasoning models on consumer-grade GPUs with limited memory. However, this would require adapting inference-optimized framework tools, such as SGLang \cite{zheng2024sglang} and vLLM \cite{kwon2023efficient}, to the Ladder architecture and we leave this research track for future work.

\begin{findingbox}
    Ladder method reaches comparable performance to QLoRA on math reasoning tasks, while being more memory-efficient and easier to train on consumer-grade GPUs.
\end{findingbox}


\begin{table*}[htbp]
\centering
\begin{tabular}{@{}lcccccc@{}}
\toprule
\textbf{Method} & \textbf{MATH-500} & \textbf{AIME24} & \textbf{AIME25} & \textbf{AMC23} & \textbf{Minerva} & \textbf{\begin{tabular}[c]{@{}c@{}}Olympiad\\ Bench\end{tabular}} \\ \midrule
Ladder@4 & 67.3 \footnotesize{$\pm$ 1.0} & 5.3 \footnotesize{$\pm$ 1.8} & 4.0 \footnotesize{$\pm$ 4.3} & \textbf{39.5} \footnotesize{$\pm$ 4.5} & 28.2 \footnotesize{$\pm$ 2.1} & 31.0 \footnotesize{$\pm$ 0.6} \\\midrule
xLadder@4 & \textbf{68.4} \footnotesize{$\pm$ 0.8} & \textbf{9.3} \footnotesize{$\pm$ 3.7} & \textbf{6.0} \footnotesize{$\pm$ 5.5} & 39.0 \footnotesize{$\pm$ 3.8} & \textbf{28.8} \footnotesize{$\pm$ 1.9} & \textbf{31.3} \footnotesize{$\pm$ 1.0} \\ \bottomrule
\end{tabular}
\caption{Results of Ladder@4 and xLadder@4 on math reasoning benchmark. The backbone model is Qwen2.5-Math-1.5B, and Ladder is connected to the first 4 layers of the backbone, while xLadder is connected to the first 2 layers with 2 additional layers. The results are averaged over 5 random seeds.}
\label{tab:math_xladder}
\end{table*}

\subsection{Experiments on xLadder}
\textbf{Setup.}
To fairly compare Ladder with xLadder, we conduct the same experiments as previously on the math reasoning benchmark
with the same depth $l=4$.
Ladder is connected to the first 4 layers of the Qwen2.5-Math-1.5B ($\mathcal{C} = \{i \rightarrow i\}_{1\le i\le 4}$,
named Ladder@4), while xLadder is connected to the first 2 layers with 2
additional layers ($\mathcal{C} = \{i \rightarrow i\}_{1\le i\le 2}$,
named xLadder@4). Therefore, both architectures have the
same number of parameters. More details on the choice of the configuration can be found in Appendix~\ref{sec:appendix-depth}.

\begin{figure}[htbp]
    \centering
    \includegraphics[width=0.9\columnwidth]{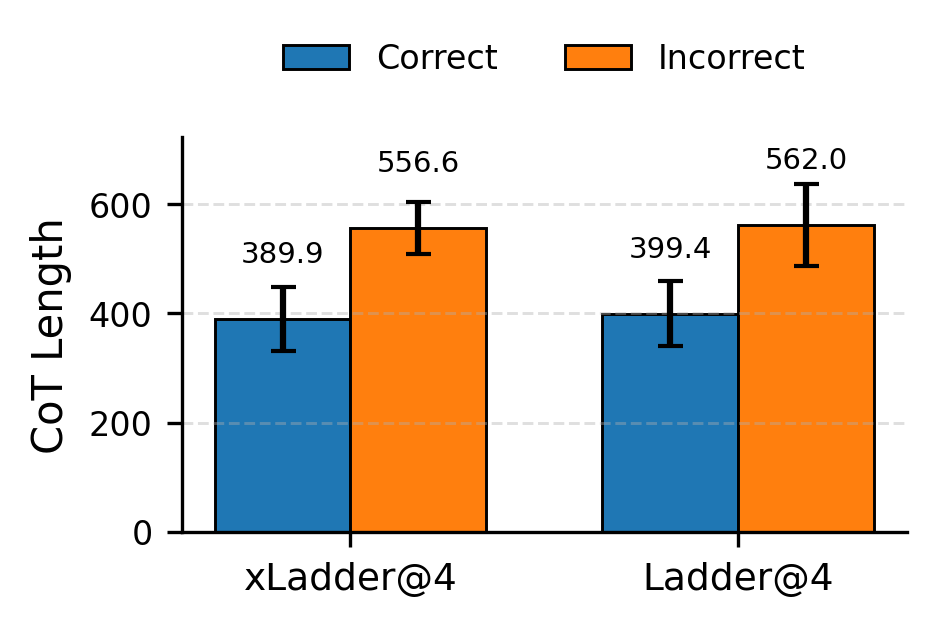}
    \caption{Reasoning average CoT length over 3 datasets for correct and incorrect answers.}
    \label{fig:cot_length}
\end{figure}

\noindent\textbf{Results.}
We report in Table~\ref{tab:math_xladder} and Figure~\ref{fig:cot_length} the results on math reasoning tasks and the average correct and incorrect CoT tokens length generated by both methods. These results show that xLadder outperforms Ladder on most datasets, while reducing the number of tokens generated in the CoT. This observation is coherent with prior work \cite{wu2025lessunderstandingchainofthoughtlength}, which shows that, in some cases, the longer the model reasons, the more likely it is to make mistakes. This suggests that the xLadder architecture may effectively extend the reasoning depth of the model without requiring additional tokens; this is the key advantage of this architecture.

We have further completed these results to compare QLoRA, Ladder and xLadder on additional benchmarks in Appendix~\ref{sec:appendix-glue-llmcritic}. More precisely, we report comparative results on CoLA and an original LLM critic dataset.
The behaviour of xLadder observed on these datasets is consistent with the previous observations.

\begin{findingbox}
    For a given number of parameters, adjusting cross-connections in xLadder to increase the depth of the forward pass might improve performance and reduce the number of tokens generated in the CoT.
\end{findingbox}



\section{Ablation Studies}
\label{sec:main-ablation}
We report next three ablations, which conclusions can be summarized as follows:
(i) optimal depth and starting position of Ladder connections depend on the task: there is no single best Ladder configuration;
(ii) the choice of Ladder width is not critical for performances, and
(iii) the weight initialization scale matters more than the initialization method.

\subsection{Ablation on Connection Patterns}
\textbf{Setup.}
First, to study the impact of depth, we progressively increase the number of
Ladder layers, assuming a simple parallel contiguous connection pattern
starting at the first layer (Figure~\ref{fig:sota_ladder}, middle):
$\mathcal{C} = \{i \rightarrow i\}_{1\le i\le l}$, progressively increasing $l=4,5,\dots$.

Second, we test by fixing the Ladder depth to 4 contiguous layers and slide it up:
$\mathcal{C} = \{i+S \rightarrow i\}_{1\le i\le 4}$ by progressively increasing $S=0, 1, 2, \dots$.
We train on math reasoning tasks and on classification tasks.
Results are given in Appendix~\ref{sec:appendix-depth}.

\noindent\textbf{Finding.}
On math tasks, deeper Ladders connected either early or late give the best results.
However, on classification benchmarks, the optimal connection pattern is different.
We conclude that there is no universally optimal connection pattern,
which is dataset and task-dependent.





\begin{figure}
    \centering
    \includegraphics[width=1.\columnwidth]{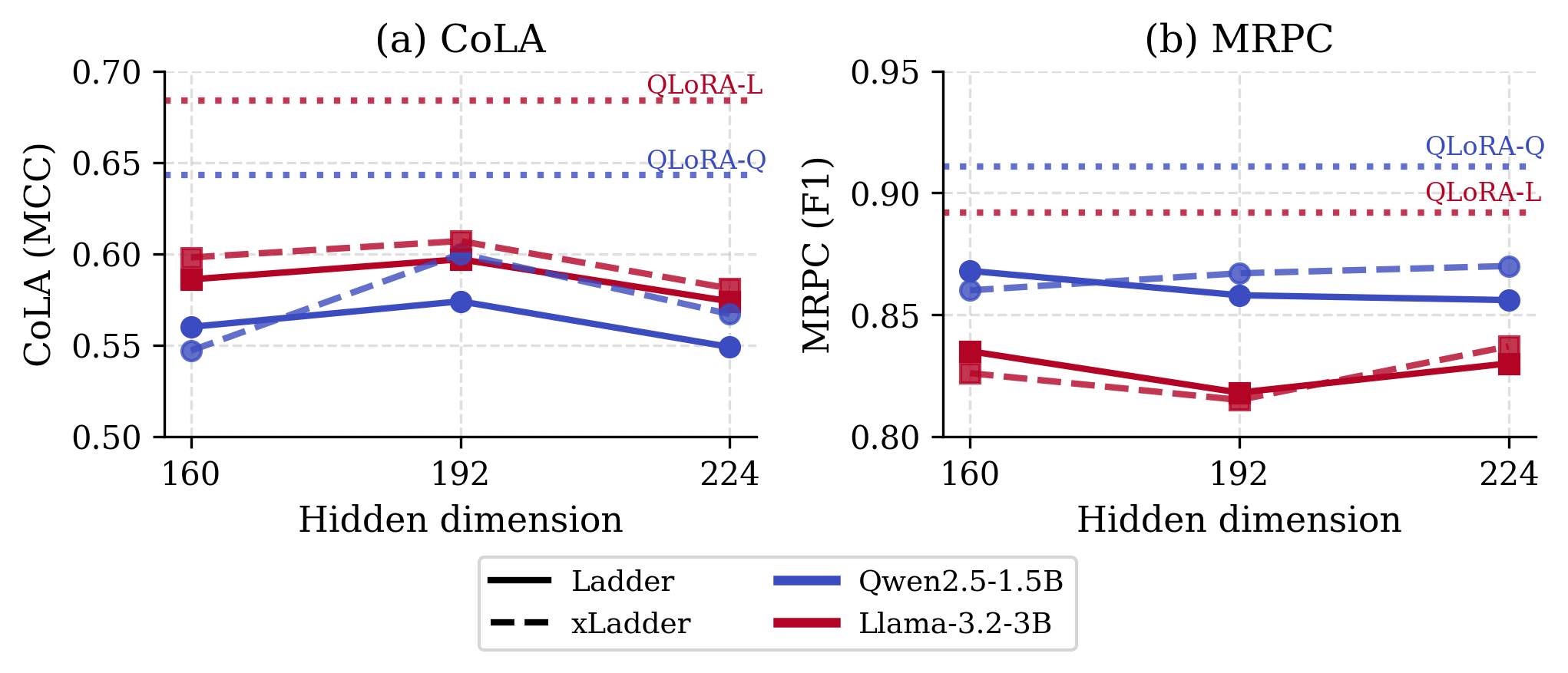}
    \caption{Width analysis on CoLA and MRPC datasets between Ladder and xLadder for Qwen-2.5-1.5B and Llama3.2-3B.}
    \label{fig:ablation_width}
\end{figure}

\subsection{Ablation on (x)Ladder Width}
\textbf{Setup.} 
We vary next the Ladder width from 160 to 224, assuming a fixed fully-connected pattern (Figure~\ref{fig:sota_ladder}, middle).
For xLadder, we further insert 8 additional layers on top.
More details on the training setup can be
found in Appendix~\ref{sec:appendix-ablation-implementation}.
Results on classification tasks are shown in Figure~\ref{fig:ablation_width}.

\noindent\textbf{Finding.}
On these tasks, we do not observe a clear consistent effect of hidden size; scores fluctuate modestly. Under the same setup, xLadder and Ladder perform comparably, with differences small relative to run-to-run variation. 

\subsection{Ablation on Weight Initialization}
\begin{figure}
    \centering
    \includegraphics[width=\columnwidth]{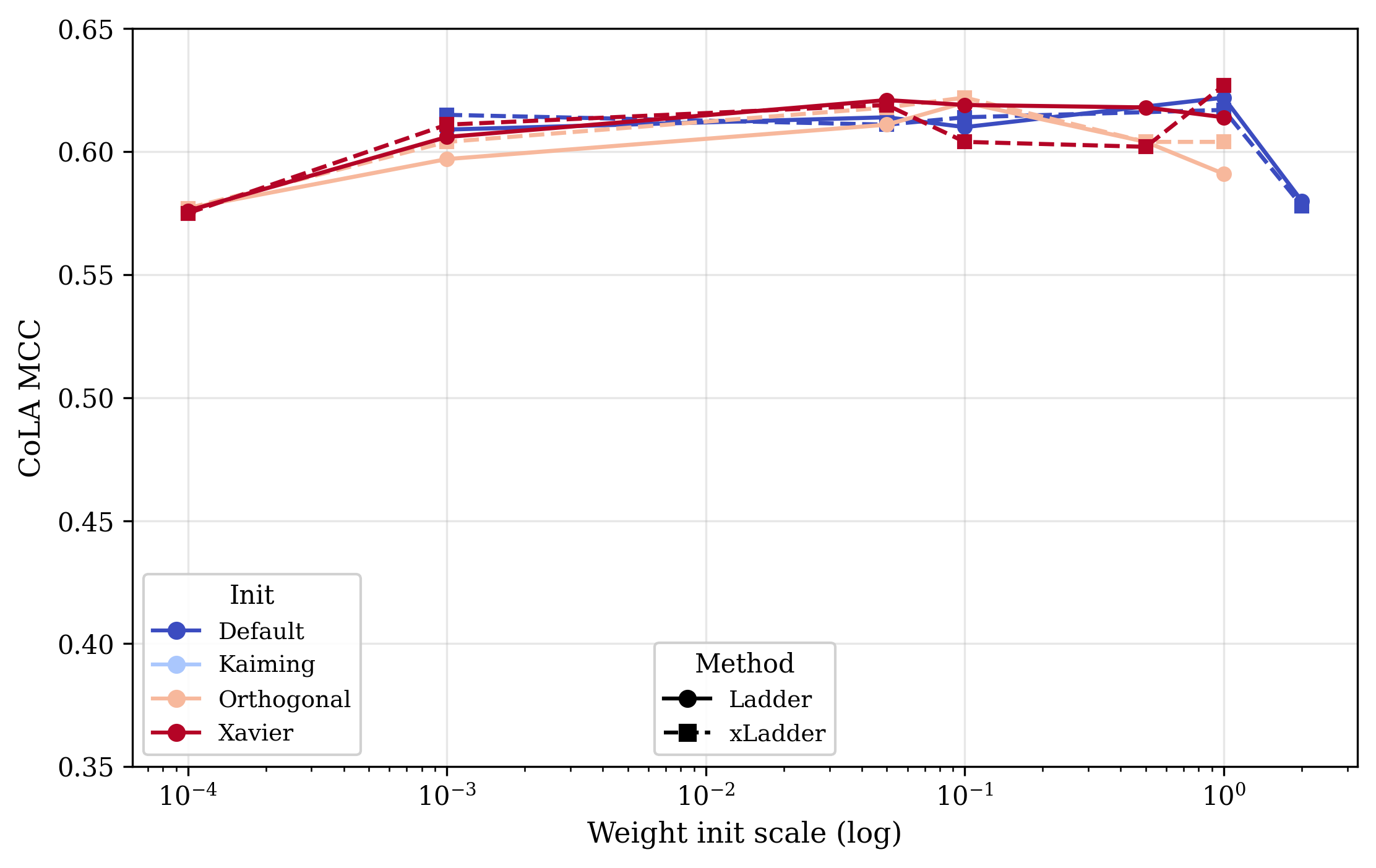}
    \caption{Weight initialization comparison on CoLA dataset.}
    \label{fig:init_plot}
\end{figure}

\textbf{Setup.} On the same classification task, we ablate weight initialization for Ladder and xLadder by
comparing different initialization strategies: uniform,
Kaiming, Xavier and Orthogonal initialization.
We also vary the weight initialization scale.
Results are shown in Figure~\ref{fig:init_plot}.

\noindent\textbf{Finding.}
Small weight scales hurt performance; progressively increasing the scale makes
accuracy rise quickly and then plateau at moderate
scales; pushing to the largest scales degrades again. Differences between
initialization methods are small (within variance), and xLadder and Ladder have the same behaviour.
Therefore, initialization scale is more important to tune than initialization variant.

\section{Conclusion}

We revisit Ladder adapters and show two key properties: 
(i) Ladder is an instance of a more flexible LLM-side-connections framework that is still largely unexplored, and
(ii) its compute-performance scaling matches QLoRA. 
Our memory analysis and reasoning benchmarks indicate Ladder is competitive while being more memory-efficient. 
Leveraging this flexibility, we introduce xLadder, which deepens reasoning per time-step,
hence reducing CoT length. 
Although broader validation across models and tasks is still needed, 
finetuning-time depth extension emerges as a new research option to enhance reasoning, 
in addition to pretraining and test-time strategies.
In future works, we plan to keep on exploring alternative connection patterns and extend the set of models and tasks.

\section*{Limitations}
\label{sec:limitations}
\textbf{Compatibility with attention optimizations.}
Our memory footprint analysis assumes vanilla attention, without optimizations
such as FlashAttention, Flex-Attention, GQA, NVMe-offloading, etc.
We acknowledge that such optimizations are important to take into account in order
to better quantify the real practical memory gain brought by Ladder; however, it is
extremely challenging to take them all into account, because there exist many such optimizations,
new ones are frequently proposed, and they might be dependent on specific hardware.
This is why we studied in the Appendix one of the most common such optimization: gradient checkpointing.
Nevertheless, we acknowledge that additional studies are required to get a better understanding
of the respective trade-offs of every optimization approach.
We let these studies for the future, noting that most of them are also complementary to Ladder,
which can therefore be also combined with them to potentially cumulate memory gains.

\noindent\textbf{Scaling laws robustness.}
Our study focuses on fine-tuning scenarios in hardware constrained settings,
and only explores a few model families. The provided scaling
laws could be more robust by validating them on more model families,
both within the same range of sizes (e.g., Qwen2.5-1.5B and Llama-3.2-1B), as well as with larger
models (e.g., Qwen2.5-14B and Phi4-14B). 
We plan to broaden the scole of our proposed scaling laws with additional models and tasks
in future works.

\noindent\textbf{Reasoning and depth.}
We illustrate Ladder's flexibility by proposing a depth-extension method, xLadder.
However, the research question about the impact of transformer depth on reasoning is
actually highly complex. For instance, naively increasing depth may induce training
instabilities, vanishing gradient... Furthermore, it is not clear what is the best compromise
between adding more layers and adding more tokens to the CoT.
Finally, many methods in the litterature explore alternative ways to increase depth, such as
Loop transformers, which loop over the same layers multiple times, latent reasoning models
(e.g., Coconut) that stays in the latent embedding domain...
In this work, we have only scratched the surface of this broad research question, and 
a solid analysis of our xLadder proposal would require a whole paper on its own,
with more experiments and comparison with the state-of-the-art.
This is also a major track of research we plan to investigate in the future.



\section*{Acknowledgments}
This project was provided with computing HPC and storage resources by GENCI at 
IDRIS thanks to the grant 2025-AD011011668R5 on the supercomputer Jean Zay's A100 
partition.
\bibliography{custom}

\clearpage
\appendix

\section{Additional Experiments}
\label{sec:appendix-glue-llmcritic}
\subsection{Classification task: LLM critic}
\label{sec:critic}

\textbf{Setup.}
We build an LLM critic dataset by sampling from the MATH-500 dataset~\cite{hendrycks2021measuringmathematicalproblemsolving}. First, several pretrained solvers (i.e., LLMs) are asked to produce a Chain-of-Thought (CoT) along with an answer for each question in the MATH-500 dataset. We keep only generations in which the \texttt{\textbackslash boxed\{\}} answer occurs within the first 2,048 tokens, then verify the correctness of the answers using Math-Verify~\cite{mathverify}. The resulting corpus is balanced (50/50 correct-incorrect) with an equal number of correct and incorrect answers, leading to a dataset of 380 training and 396 test samples, with an average length of 600 tokens. This dataset will be freely available on HuggingFace. 

For the critic itself, we fine-tune Qwen2.5-0.5B and Qwen2.5-3B backbones with the Ladder and QLoRA methods on 3 epochs with a learning rate of 5E-4. These hyperparameters were selected based on the convergence of the training curve. The task is binary: predict whether the input CoT is correct or incorrect. 

The prompt used for training and evaluation is provided in Appendix~\ref{sec:appendix-llm_critic}.




\noindent\textbf{Results.}
Table~\ref{tab:critic} shows the results of our LLM critic on the test set. Both fine-tuning methods significantly improve the near-random performance of the base models to over 80\% accuracy. As both methods achieved similar results, we can conclude that training data quality and implementation constraints (i.e., limited GPUs) are more important than the fine-tuning method itself. 

\begin{table}[htbp]
\centering
\resizebox{.85\width}{!}{%
\begin{tabular}{@{}l c c@{}}
\toprule
    \textbf{Model} & \textbf{Method} & \textbf{\% Acc} \\
    \midrule
    Qwen2.5-0.5B & Base & 59.0 \\
    Qwen2.5-0.5B & Ladder &  {\bf 81.8} \\
    Qwen2.5-0.5B & xLadder & 79.0  \\
    Qwen2.5-0.5B & QLoRA & 79.0 \\
    \midrule
    Qwen2.5-3B & Base & 56.0 \\
    Qwen2.5-3B & Ladder & 81.0 \\
    Qwen2.5-3B & xLadder & 82.0 \\
    Qwen2.5-3B & QLoRA & {\bf 82.8} \\
    \bottomrule
\end{tabular}%
}
\caption{LLM critic results on 396 test samples. 95\% Wald-Confidence interval is $\pm 3.9\%$.}
\label{tab:critic}
\end{table}

LLMs are increasingly used to assess the quality of the generation of other LLMs, e.g., LLM-as-a-Judge tasks~\cite{zheng-etal-2023-judging}, reward modeling for RL fine-tuning~\cite{lee-etal-2024-rlaif, bai2022constitutionalaiharmlessnessai}, and grading critics~\cite{ke2024critiquellminformativecritiquegeneration}.
Our critic fits this trend, it provides a fast, and 82\% accurate filter that screens out most flawed CoTs before passing the higher confidence ones to a more expensive process such as human evaluation or RL-based fine-tuning.

The critic has a residual error of 18\% and the modest training dataset size leaves room for robustness improvement under distribution shift, such as the use of more diverse pretrained LLMs or the use of more complex prompts.



\subsection{Classification task: CoLA}
\label{sec:appendix-cola}
\textbf{Setup.}
In the GLUE benchmark, we focus on the CoLA dataset, which is a binary classification task on the grammatical acceptability of sentences. We use the same parameters provided in prior LST work \cite{sung2022lst}. 

We reimplement the architecture as the original code was not compatible with recent versions of the Python libraries. The original LST architecture is implemented with a side network that is connected to every layer of the backbone model, and it is initialized by pruning the backbone model weights. Additional experiments with a slightly different and simpler architecture are also conducted, especially with an uniform initialization and an unweighted gated sum $\alpha$ between the ladder and backbone activations, instead of a weighted sum as in the original LST architecture.

\begin{table}[htbp]
\centering
\resizebox{.9\width}{!}{%
    \begin{tabular}{@{}lccc@{}}
    \toprule
    \textbf{Model} & \textbf{Method} & \textbf{Reproduction} & \textbf{Original}\\ \midrule
        OPT1.3B & QLoRA & 0.630 & {\it 0.621 $\pm$ 0.023} \\
        OPT1.3B & LST & 0.579 & {\it 0.595 $\pm$ 0.031} \\ \midrule
        OPT2.7B & QLoRA & 0.656 & {\it 0.637 $\pm$ 0.026} \\
        OPT2.7B & LST & 0.592 & {\it 0.607 $\pm$ 0.035} \\ \midrule
        OPT6.7B & QLoRA & 0.665 & {\it 0.643 $\pm$ 0.028} \\
        OPT6.7B & LST & 0.596 & {\it unreported} \\ \bottomrule
    \end{tabular}%
}
\caption{Reproduction of the results reported in~\cite{sung2022lst} on the CoLA benchmark with our Ladder codebase. The metric used is MCC.}
\label{tab:glue_res}
\end{table}

For reproduction purposes, the backbone models used are OPT family models \cite{zhang2022optopenpretrainedtransformer}, which are 1.3B, 2.7B and 6.7B parameters. Additional newer models such as Llama3.2-3B and Qwen2.5-1.5B are also used to compare the Ladder architecture with the QLoRA method.

We trained on 7 epochs with early stopping with a learning rate of 1E-5, the AdamW optimizer and a cosine learning-rate schedule. All the results are 4 seeds average. The is performed on a single ADA RTX5000 32GB GPU.


\noindent\textbf{Results.}
Re-running the original LST experiments on the CoLA benchmark, we notice a difference in the results compared to the original paper as shown in Table~\ref{tab:glue_res}. This difference is likely due to the use of different library versions and GPU kernels, which can lead to slight variations in the results. 
However, the differences observed on the Ladder models are all within the range of variability reported in the paper, which validates our reproduction and open-source codebase. 

\begin{table}[htbp]
\centering
\resizebox{.9\width}{!}{%
    \begin{tabular}{@{}lcc@{}}
    \toprule
    \textbf{Model} & \textbf{Method} & \textbf{MCC}\\
\midrule
        Llama3.2-3B & QLoRA     & \textbf{0.69} \\
        Llama3.2-3B & Ladder &  0.61 \\
        Llama3.2-3B & xLadder &  \underline{0.64} \\
\midrule
        Qwen2.5-1.5B & QLoRA & \textbf{0.62} \\
        Qwen2.5-1.5B & Ladder & 0.58 \\
        Qwen2.5-1.5B & xLadder & \underline{0.59} \\
\bottomrule
    \end{tabular}%
}
    \caption{Additional results on CoLA with newer models and a simpler ladder: uniform initialization, sum gates between ladder and backbone $\alpha=0.5$. The Wald 95\% Confidence Interval is $\pm 0.03$.}
\label{tab:glue_res2}
\end{table}

Furthermore, comparing the Ladder with the xLadder, we observe that the xLadder architecture seems to consistently outperform the Ladder architecture as shown in Table~\ref{tab:glue_res2}. This suggests that the additional layers and connections in the xLadder architecture provide a performance boost over the standard Ladder architecture. Finally, when comparing the Ladder method with QLoRA, we observe that the ladder-based methods achieve slightly lower results than QLoRA on the CoLA classification task. This suggests that initialization by pruning  tends to give slightly better results.

\section{Additional Ablation Studies}
\subsection{More Details on Depth \& Starting Placement}
\label{sec:appendix-depth}
\subsubsection{Math Reasoning Task}
\label{sec:appendix-math-depth}
\textbf{Setup.}
We train Qwen2.5-Math-1.5B on the open-s1 dataset \cite{openr1,muennighoff2025s1simpletesttimescaling} (a filtered 52k version of s1) for 1 epoch, with learning rate 2E-4 and batch size 2, on a single NVIDIA A100-80GB. Then, we fixed the Ladder depth to 4 layers, and vary the starting index along the backbone model. We report the accuracy on the MATH-500 dataset. QLoRA is trained with standard hyperparameters $r=$16, $\alpha=$32 and dropout=0.005.

\begin{figure}[htbp]
    \centering
    \includegraphics[width=0.8\columnwidth]{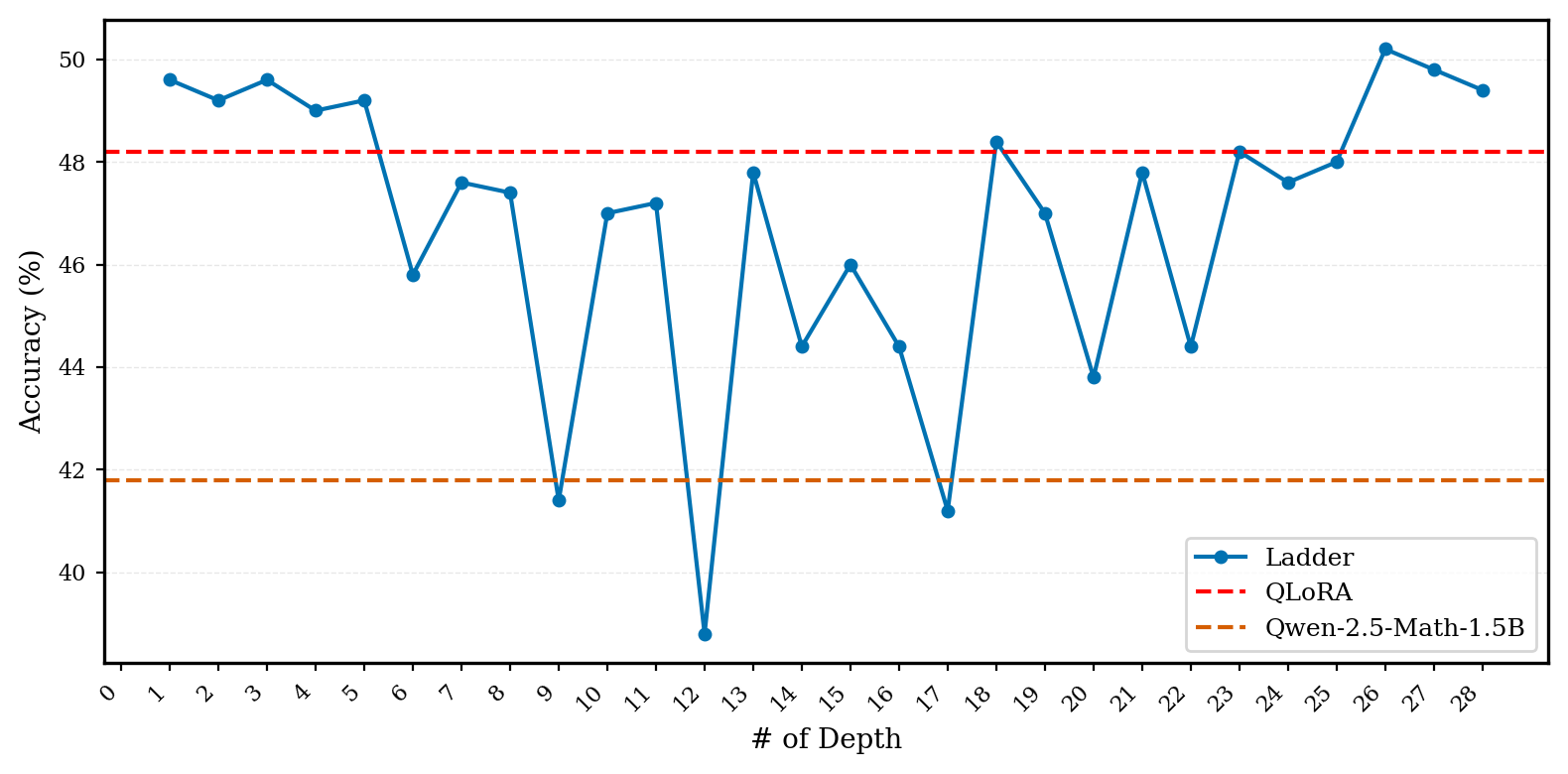}
    \caption{Accuracy of different number of layers connected on MATH500. Red dashed line represents the QLoRA baseline. Orange dashed line represents the base model accuracy.}
    \label{fig:growth_opens1}
\end{figure}

\noindent\textbf{Results.}
Figure \ref{fig:growth_opens1} confirms the U-shape accuracy curve. This shows that either fully connected Ladder or very few connected layers yield the best results. The noisiness of the curve is due to the limited one training seed run, and its sensitivity to random factors such as training seed or batch samples.

\begin{figure*}[htbp]
    \centering
    \includegraphics{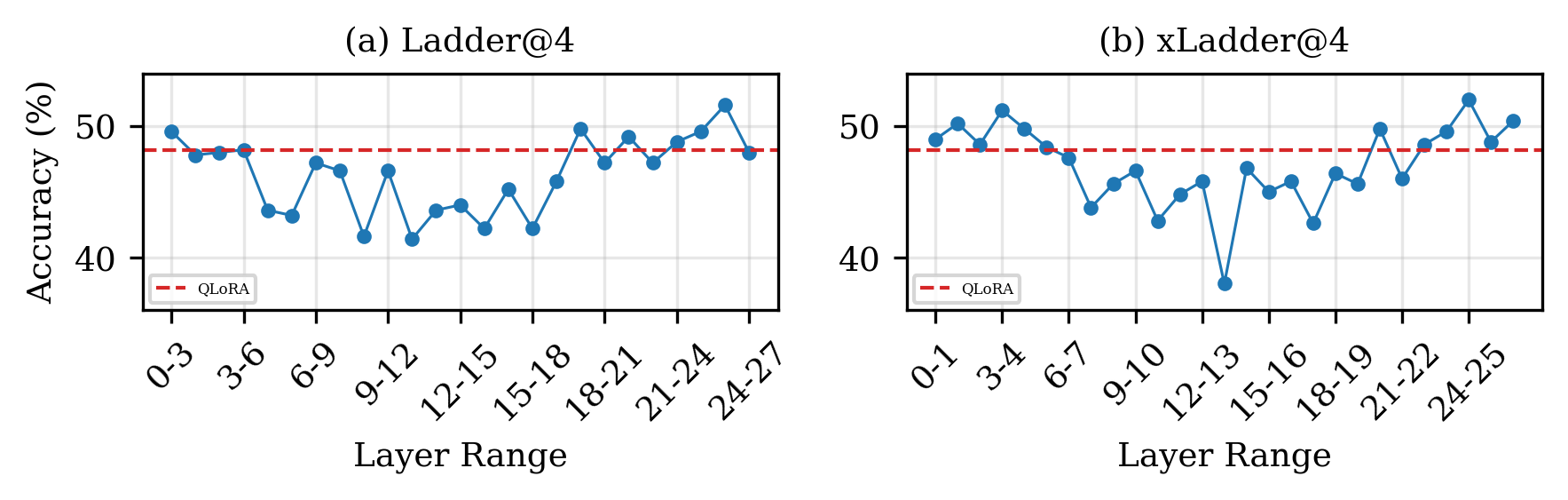}
    \caption{Starting position of the connected layers on MATH-500. The red dashed line represents the QLoRA baseline. On the left, (a) is Ladder for a depth of 4, and on the right, (b) xLadder for a depth of 4 including 2 additional layers.}
    \label{fig:placement_opens1}
\end{figure*}

This U-shape trend remains when varying the starting position of the connected layers as shown in Figure \ref{fig:placement_opens1}. The xLadder seems to be more sensitive to the starting position, which suggests that the additional layers and connections in the xLadder architecture may amplify the effects of starting position, making it more important to choose an optimal starting point for maximum performance.

\subsubsection{GLUE Classification Tasks}
\label{sec:appendix-glue-depth}
\textbf{Setup.}
We train Qwen2.5-1.5B and Llama3.2-3B on the CoLA and MRPC datasets from the GLUE benchmark \cite{wang-etal-2018-glue} for 7 epochs, with learning rate 1E-4 and a cosine learning rate schedule. All the results are 4 seeds average. The training is performed on a single ADA RTX5000 32GB GPU. We report the MCC (Mathews Correlation Coefficient) for CoLA.
Ladder is fixed to depth 5.

\begin{figure}
    \centering
    \includegraphics[width=\columnwidth]{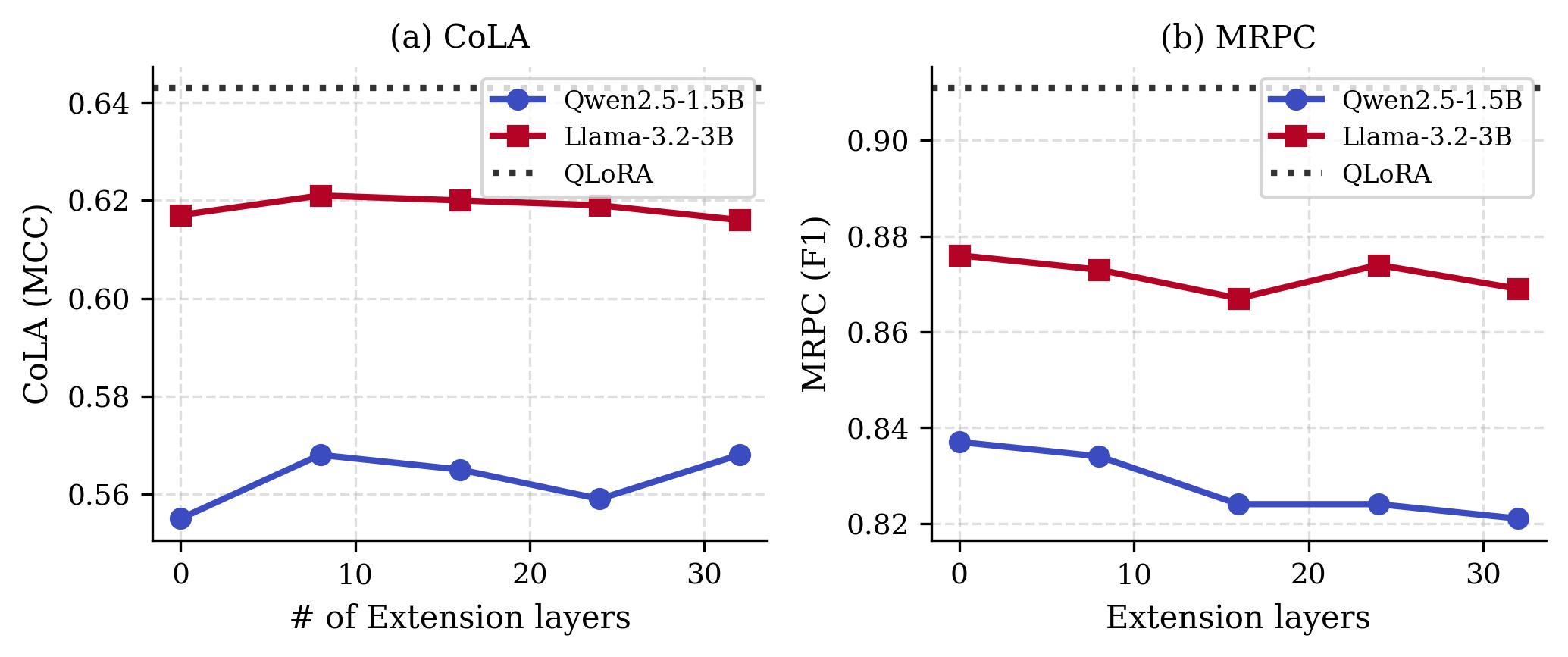}
    \caption{Performance on CoLA and MRPC on the number of extended layers for the xLadder architecture. Ladder depth is fixed to 5.}
    \label{fig:extension}
\end{figure}

\noindent\textbf{Results.}
Figure \ref{fig:extension} shows that increasing the depth of non-connected layers on the xLadder architecture does not significantly impact the performance on both CoLA and MRPC datasets. 

\begin{figure}
    \centering
    \includegraphics[width=\columnwidth]{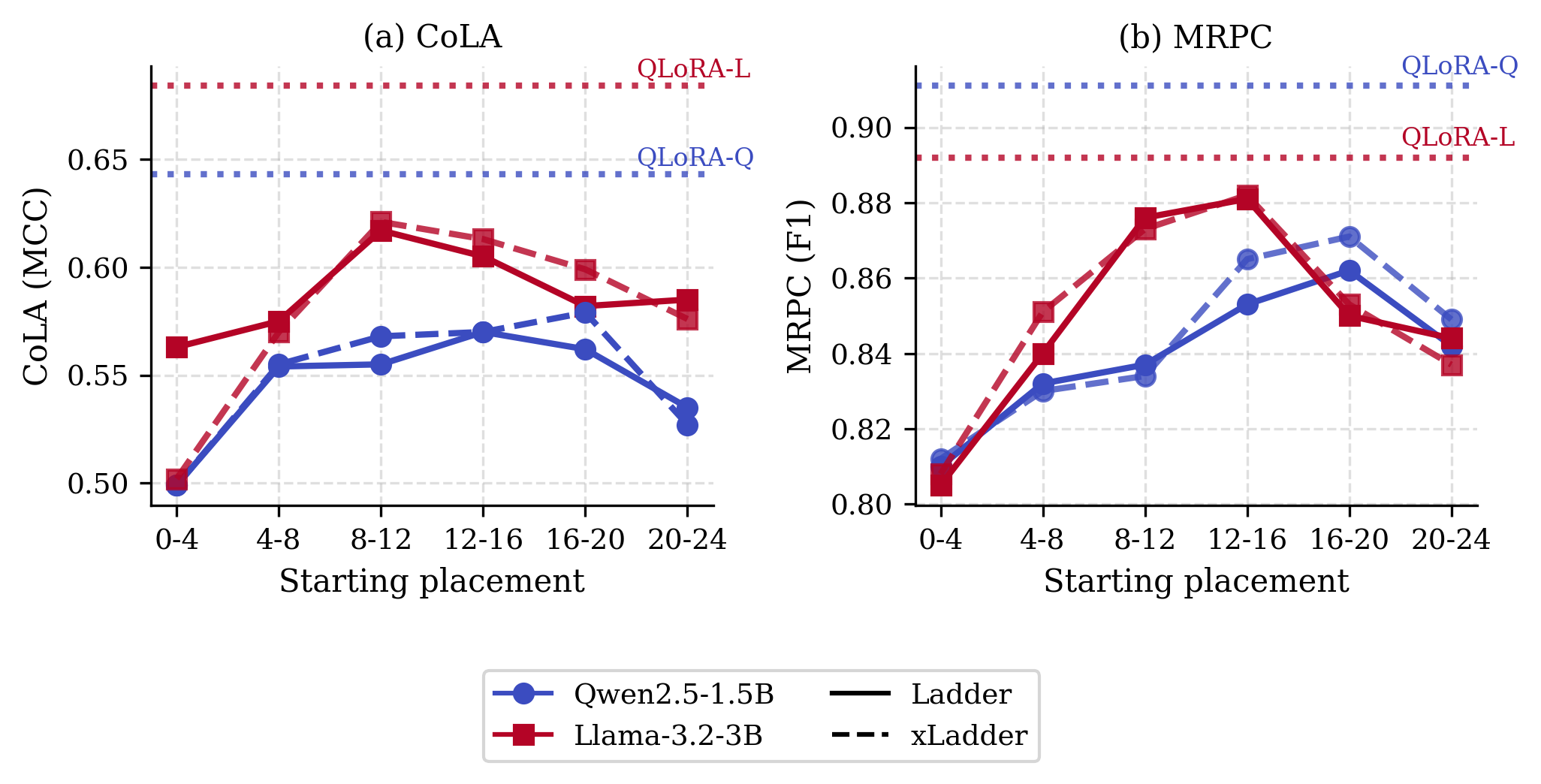}
    \caption{Starting position of the connected layers on CoLA and MRPC datasets between Ladder and xLadder for Qwen-2.5-1.5B and Llama3.2-3B. Ladder is fixed to depth 5, and xLadder to depth 5 with 8 additional layers.}
    \label{fig:placement_glue}
\end{figure}

Figure \ref{fig:placement_glue} shows the starting position of the connected layers for both Ladder and xLadder architectures. The trend is different from the one observed on math reasoning tasks, as the optimal starting position seems to be more centered in the middle layers. This suggests that the optimal starting position for connected layers may vary depending on the specific task and dataset, and that different tasks may require different strategies for selecting the starting position of connected layers in Ladder-based architectures.

\subsection{Implementation Details on Width and Initialization Ablations}
\label{sec:appendix-ablation-implementation}
The same setup is used for both experiments. We use Qwen2.5-1.5B and Llama3.2-3B backbones and train on the CoLA and MRPC datasets from the GLUE benchmark \cite{wang-etal-2018-glue} for 7 epochs, with learning rate 1E-4 and a cosine learning rate schedule. All the results are 4 seeds average. The training is performed on a single ADA RTX5000 32GB GPU. We report the MCC (Mathews Correlation Coefficient) and F1 scores for CoLA and MRPC, respectively. 

For QLoRA, we use standard hyperparameters $r=$16, $\alpha=$32 and dropout=0.005.

For Ladder and xLadder, we use a fully connected architecture and a random uniform initialization;
for xLadder, we further add 8 non connected layers on top.



\section{Details on xLadder Architecture}
\label{sec:appendix-xladder}
\subsection{Definition}
\begin{figure}[htbp]
    \centering
    \includegraphics[width=\columnwidth]{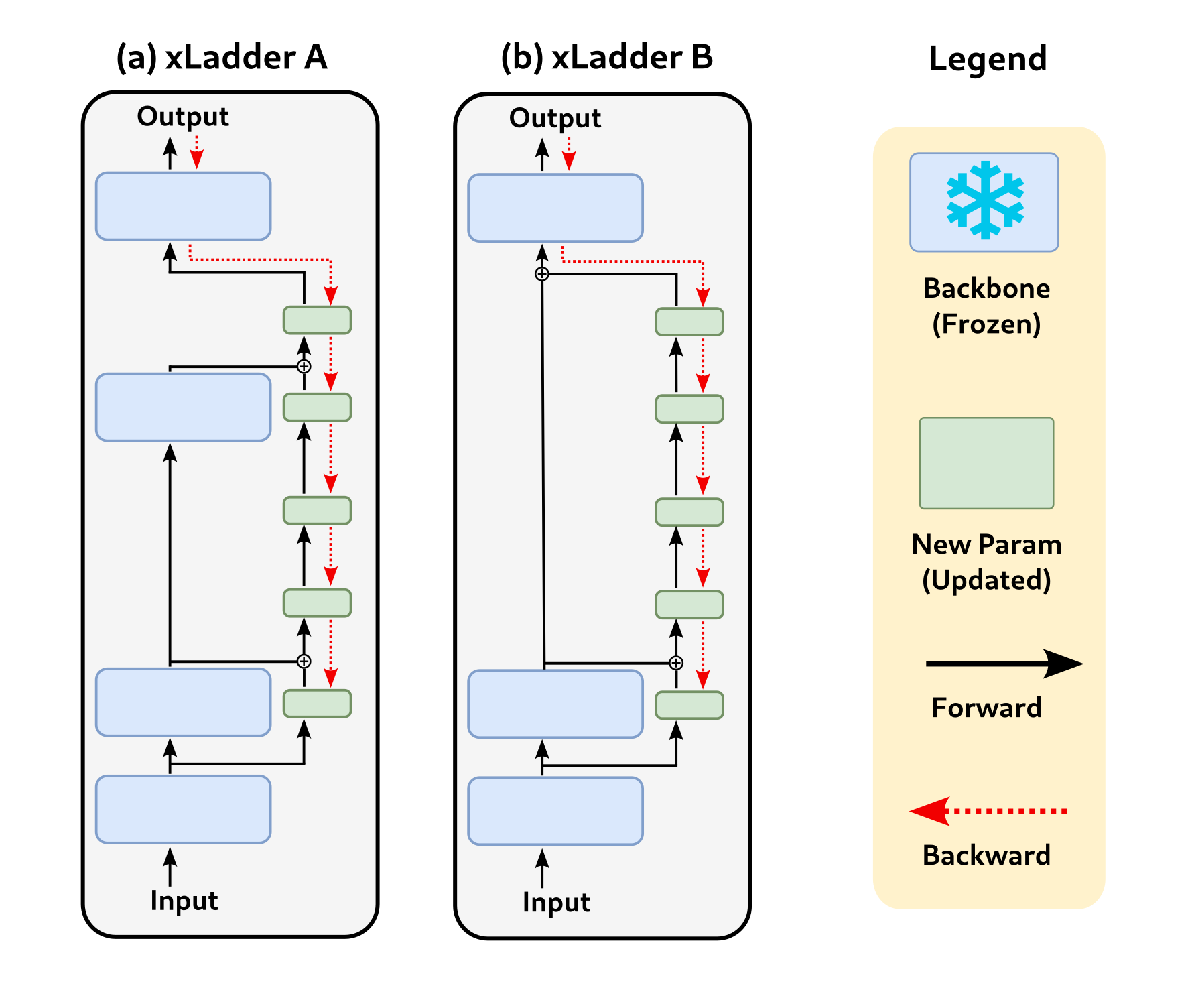}
    \caption{Overview of the xLadder architecture.}
    \label{fig:xladder}
\end{figure}

As defined in Section \ref{sec:methodology}, the xLadder architecture is an extension of the Ladder architecture that introduces additional layers to enhance reasoning depth while maintaining the efficiency of the original Ladder design. The xLadder architecture is designed to improve token efficiency in reasoning tasks by introducing a new concept called \textit{free layers}, i.e., non-connected layers. 
Figure \ref{fig:xladder} illustrates the xLadder architecture, which consists of a frozen backbone model \( f \) and a side model \( g \) built on top of the backbone. The side model is composed of a series of \textit{connected layers} \( k \) that are directly connected to the backbone model and a series of \textit{free layers} \( m \) that are not connected to the backbone model. 

We created two variants of the xLadder architecture where the difference comes from the last layer: in the first variant (Figure \ref{fig:xladder} (a)), the last layer of the backbone model is always connected to the side model before it goes to the backbone head , while in the second variant (Figure \ref{fig:xladder} (b)), the last layer of the side model is not connected to the backbone's last layer. The first variant inspired by the original LST architecture \cite{sung2022lst}, while the second variant is inspired by the QST architecture \cite{zhang-etal-2024-quantized}. The choice of the last layer connection depends on the specific task and the desired reasoning depth.

\begin{figure}[htbp]
    \centering
    \includegraphics[width=\columnwidth]{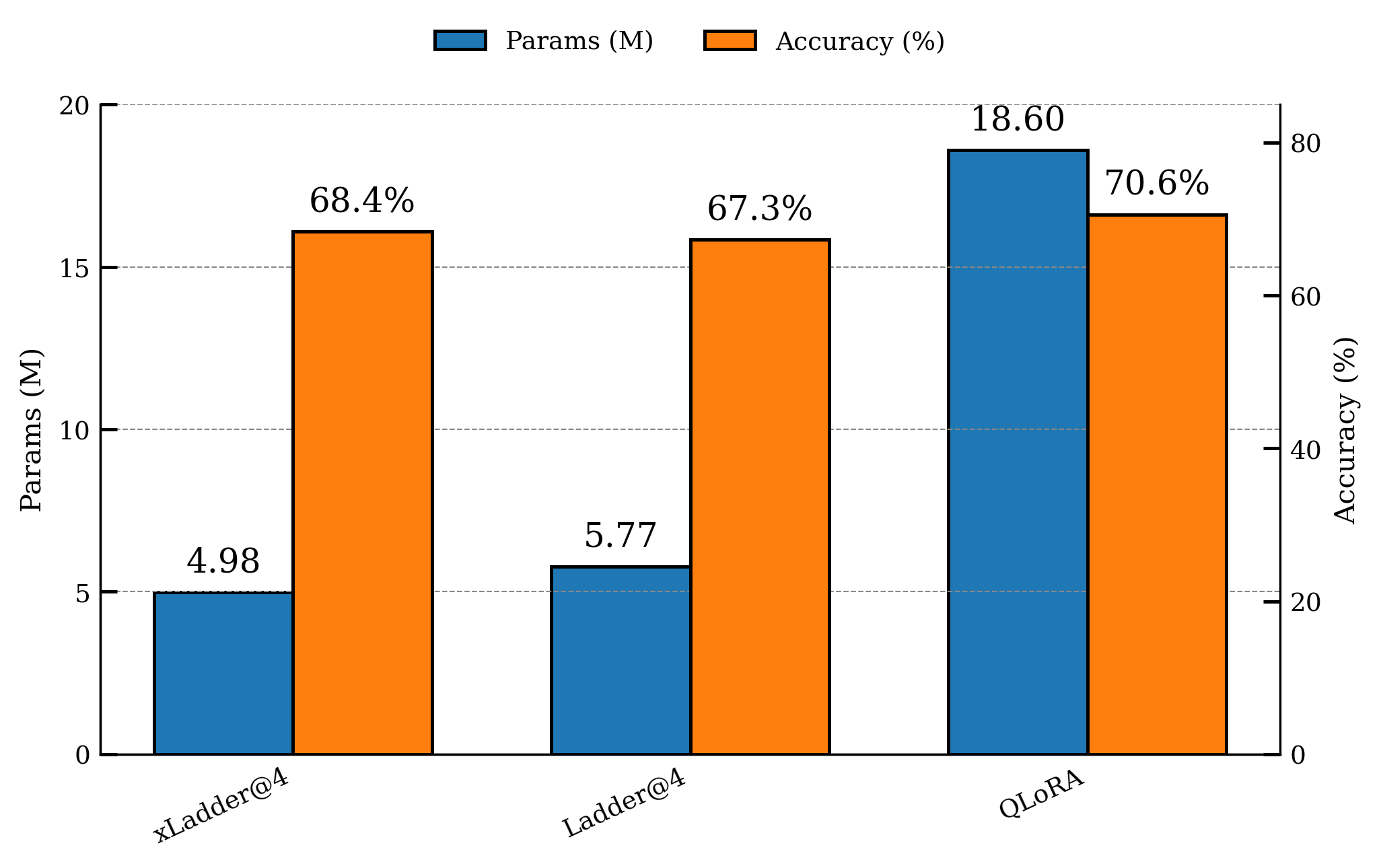}
    \caption{Avg@5 accuracy on MATH-500 for xLadder@4, Ladder@4 and QLoRA.
    The backbone model is Qwen2.5-Math-1.5B. QLoRA requires more than 3 times the number of parameters to achieve similar results.}
    \label{fig:xlad_vs_lad_vs_qlora}
\end{figure}

The connected layers are responsible for processing the input tokens and generating intermediate representations, while the free layers can be engaged to enhance the model's reasoning capabilities without requiring additional tokens. Therefore, this fine-tuning technique is memory efficient and allows for effective training without excessive resource consumption, while still maintaining high performance comparable to QLoRA, as shown in Figure \ref{fig:xlad_vs_lad_vs_qlora}.



\subsection{Investigating Depth Extension}
Depth (number of layers) limits how many sequential reasoning steps a Transformer can apply to a fixed input. Simply adding layers often harms optimization and memory (e.g., vanishing gradients)~\cite{loop1}. A common workaround is chain-of-thought (CoT): the model appends intermediate tokens and reprocesses them, which boosts performance but increases latency and memory; some works also loop selected layers at test time~\cite{li2025skiplayerloopit}. 

We propose a Ladder that extends effective depth without incurring full backprop through the added layers. With $N$ backbone layers, we connect the side network to $N/2$ backbone layers and add $N/2$ extra, non-connected layers. Thus gradients traverse a standard $N$-layer stack, while the forward pass spans $2N$ layers, therefore doubling sequential computation at training-time memory comparable to $N$. This aligns with the GLM-4.5 observation that \textit{"deeper models exhibit better reasoning capacity."}\footnote{\url{https://z.ai/blog/glm-4.5}}

\subsection{xLadder Configuration}
We simplify the original Ladder architecture \cite{sung2022lst,zhang-etal-2024-quantized} in four ways: (i) we do not connect all layers to the side model, and (ii) we fix the width of the connected layers to a constant, (iii) we replace the different projection methods to a linear projection, and (iv) we initialize the side model with a uniform initialization instead of using the backbone model weights (i.e., pruning method).

Each connected layer \( k \), which are smaller width-wise by design, is connected to the backbone model \( f \) via a linear projection, which allows the model to leverage the pre-trained knowledge of the backbone while still being able to adapt to specific tasks. Compared to current ladder approaches, this simplified architecture is more flexible and efficient since it does not require all layers to be trainable.

The additional free layers \( m \) are designed to be the same as the connected layers, but without the backbone connection. Therefore, the computational complexity of these layers is \( \mathcal{O}(d_s^2) \), where \( d_s \) is the dimension of the hidden states in the side model. 

The side model \( g \) last layer is summed with the output of the backbone model \( f \) to form the final output. As stated in QST \cite{zhang-etal-2024-quantized}, this design is effective against initialisation drift.

\section{Scaling Law Implementation Details}
\subsection{Test Loss Scaling Hyperparameters Details}
\label{sec:appendix-scaling-hyperparam}
For the scaling laws experiments in test loss scaling, we train for one epoch on each $D_f = \{50k, 100k, 200k, 300k, 400k\}$ sample with a random seed of 39. The best model is selected based on the validation loss after each epoch on a subset of EvoLM dataset \cite{qi2025evolmsearchlostlanguage}. To speed up the training, different batch sizes are used as stated in Table~\ref{tab:scaling_batchsize} while learning rate is kept constant at 2E-5 with a cosine scheduler with warmup ratio of 0.1. The optmizer is 8-bit AdamW with weight decay of 0.01.
The sequence length is set to the maximum length of the dataset, which is 1,839 tokens. QLoRA configuration is set to 4-bit quantization with $r=16$ and $\alpha=32$ on all linear layers.

All those runs can be performed on a single NVIDIA A100 GPU with 80GB of memory, with training taking approximately an hour for the Ladder and two hours for the QLoRA model on the 100k dataset sample.

\begin{table}[htbp]
\centering
\resizebox{.85\width}{!}{%
\begin{tabular}{@{}lccc@{}}
\toprule
\textbf{Model} & \textbf{Method} & \textbf{\begin{tabular}[c]{@{}c@{}}Micro \\ Batch Size\end{tabular}} & \textbf{\begin{tabular}[c]{@{}c@{}}Gradient\\ Acc\end{tabular}} \\ \midrule
    Qwen2.5-1.5B & Ladder & 8 & 3 \\
    Qwen2.5-1.5B & QLoRA & 4 & 5 \\ \midrule    
    Qwen2.5-7B & Ladder & 4 & 3 \\
    Qwen2.5-7B & QLoRA & 2 & 5 \\
\bottomrule
\end{tabular}%
}
\caption{Hyperparameters for the test loss scaling experiments. The effective batch size is the product of the micro batch size and the gradient accumulation steps.}
\label{tab:scaling_batchsize}
\end{table}

\subsection{Test Loss Scaling Ladder Configuration Details}
\label{sec:appendix-scaling-ladder-config}
\begin{figure}[htbp]
\centering
\includegraphics[width=\columnwidth]{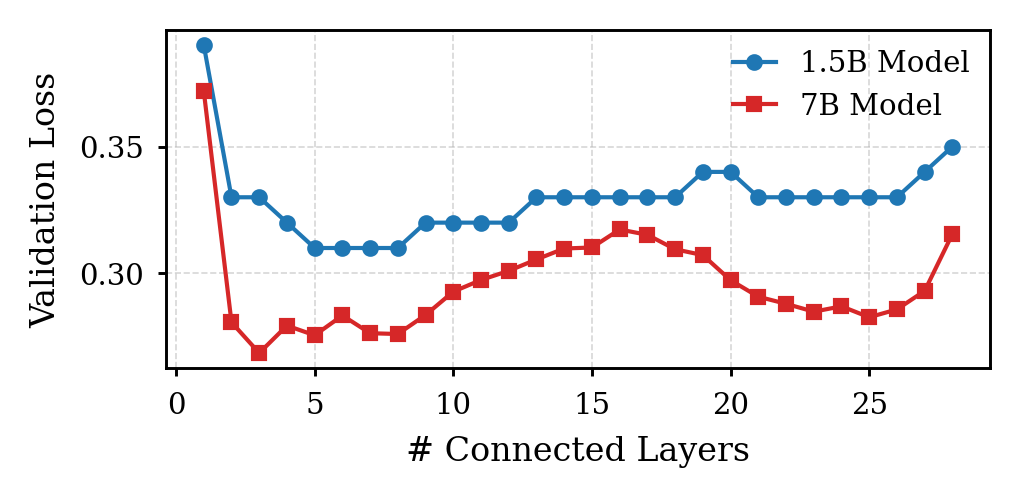}
\caption{Test loss for different Ladder depth-configurations on the 10k dataset sample. The depth goes from 1 to $L=28$, where $L$ is the number of layers in the backbone model.}
\label{fig:growth_k_short}
\end{figure}

To configure the Ladder architecture the most optimal way, we perform a hyperparameter search on the depth while keeping constant the width to 256. The depth is defined as the number of layers in the Ladder architecture. The search is performed on a 10k dataset sample, and the results are shown in Figure~\ref{fig:growth_k_short}. The best depth configuration is found to be around 6 layers for 1.5B model, while 4 layers for 7B model. We decide to use those depth configurations for the Ladder architecture in the scaling laws experiments.

The overall configuration of the Ladder is that the side network is uniformely initialized, as well as its linear projection, with unweighted gated sum (i.e., no more $\alpha$) as shown in Figure~\ref{fig:sota_ladder}.

\subsection{Downstream Tasks Hyperparameters Details}
\label{sec:appendix-downstream-hyperparam}
On downstream tasks, we trained on multiple epochs on two datasets: 7 epochs on CoLA, and 4 epochs on QQP. The best model is selected based on the validation loss computed on each epoch. Ladder use a batch size of 4, while QLoRA could not learn properly with small batch size, so we use a batch size of 32. Their learning rate is set to 1E-5 with a cosine scheduler. The optimizer used is normal AdamW.

The QLoRA configuration is the same as previously, with 4-bit quantization and $r=16$ and $\alpha=32$ on all linear layers. The Ladder architecture is a fully connected ladder on each layer of the backbone model, with a uniform initialization of the side network and its linear projection.

\subsection{Downstream Tasks Scaling Laws Equations Fit}
\label{sec:appendix-scaling-fit}



Table \ref{tab:beta-nonpooled} shows the fitted scaling law parameters for the Ladder and QLoRA methods on the downstream task experiments. We report solely the exponent $\beta$ of the error-compute power law stated in Section \ref{sec:scaling_acc} Eq.\eqref{eq:scaling_acc}, fitted using the Scipy's \texttt{curve\_fit} function. The confidence intervals are computed using a bootstrap method with 95\% confidence level. 

\begin{table}[htbp]
\centering
\begin{tabular}{@{}l c c@{}}
\toprule
\textbf{Model} & \textbf{Ladder $\beta$} & \textbf{QLoRA $\beta$} \\\midrule
Qwen-1.5B & 0.15 \footnotesize{[-0.1, 0.3]} & 0.45 \footnotesize{[0.2, 0.7]}  \\
Qwen-3B   & 0.15 \footnotesize{[-0.1, 0.3]} & 0.33 \footnotesize{[0.1, 0.6]}  \\ \midrule
OPT-1.3B  & 0.12 \footnotesize{[-0.4, 0.6]} & 0.26 \footnotesize{[0.0, 0.5]}  \\
OPT-2.7B  & 0.13 \footnotesize{[-0.2, 0.4]} & 0.24 \footnotesize{[-0.3, 1.3]}   \\
OPT-6.7B  & 0.13 \footnotesize{[-0.1, 0.3]} & 0.22 \footnotesize{[0.3, 1.2]}   \\ \midrule
Llama-1B  & 0.09 \footnotesize{[-0.8, 1.0]} & 0.45 \footnotesize{[0.2, 0.7]} \\
Llama-3B  & 0.1 \footnotesize{[-0.3, 0.5]} & 0.13 \footnotesize{[-0.5, 0.7]}  \\
\bottomrule
\end{tabular}
\caption{Scaling law parameter for QLoRA and Ladder methods on downstream tasks. 
         The 95\% Bootstrap CI is reported.}
\label{tab:beta-nonpooled}
\end{table}

\begin{figure}[htbp]
\centering
\includegraphics[width=\columnwidth]{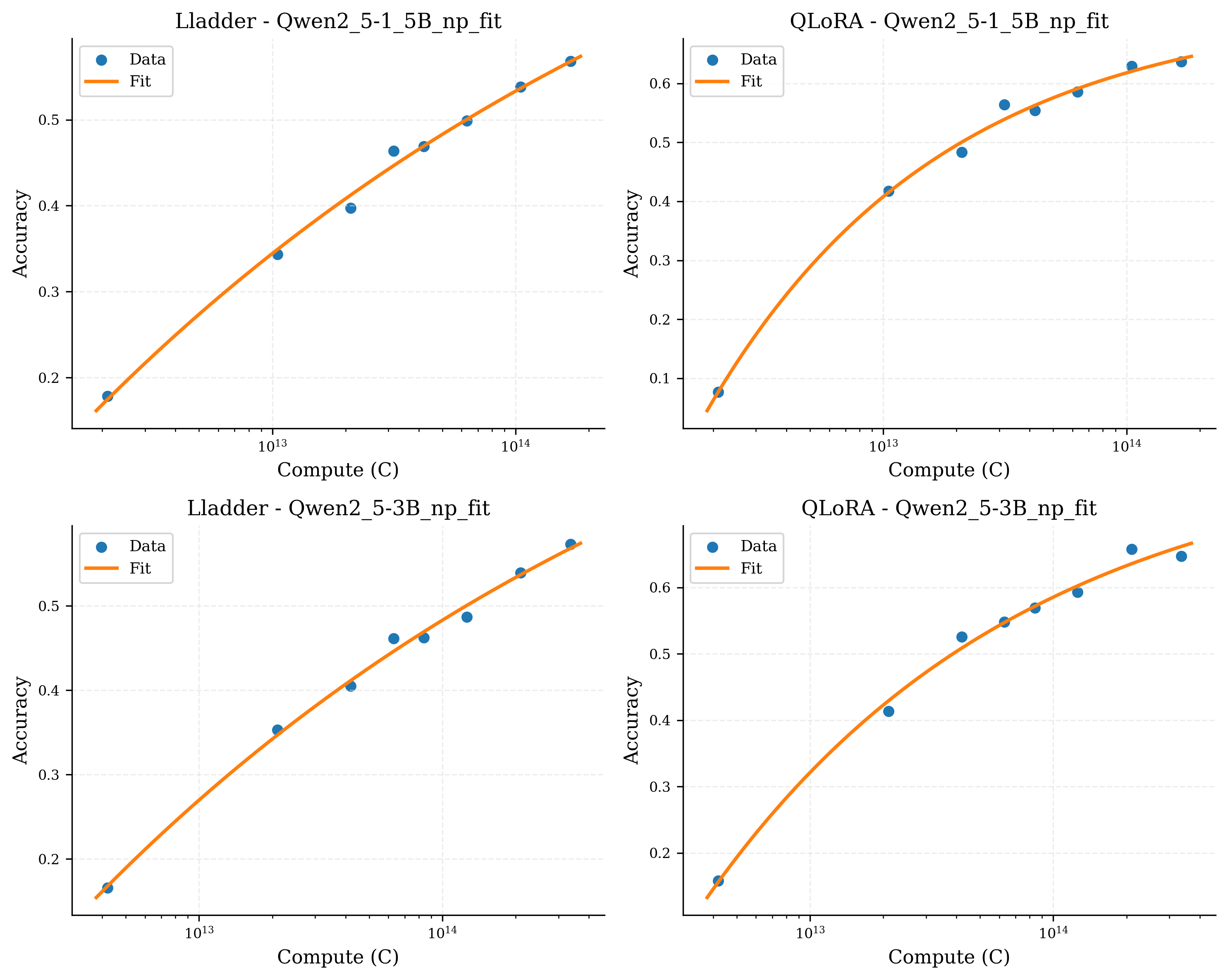}
\caption{Downstream scaling law for Qwen series models comparing Ladder and QLoRA methods.}
\label{fig:qwen_scalacc}
\end{figure}

\begin{figure}[htbp]
\centering
\includegraphics[width=\columnwidth]{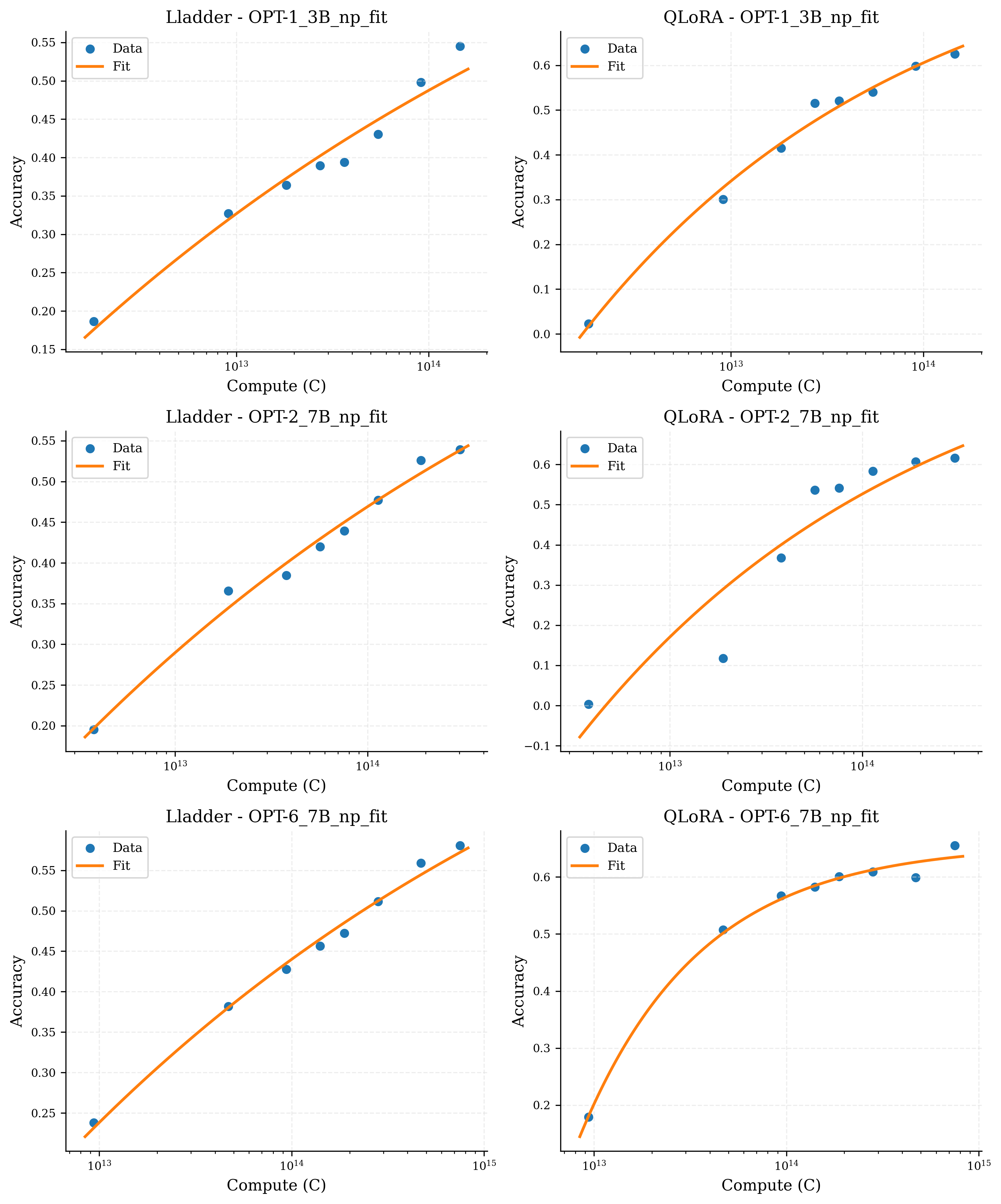}
\caption{Downstream scaling law for OPT series models comparing Ladder and QLoRA methods.}
\label{fig:opt_scalacc}
\end{figure}

\begin{figure}[htbp]
\centering
\includegraphics[width=\columnwidth]{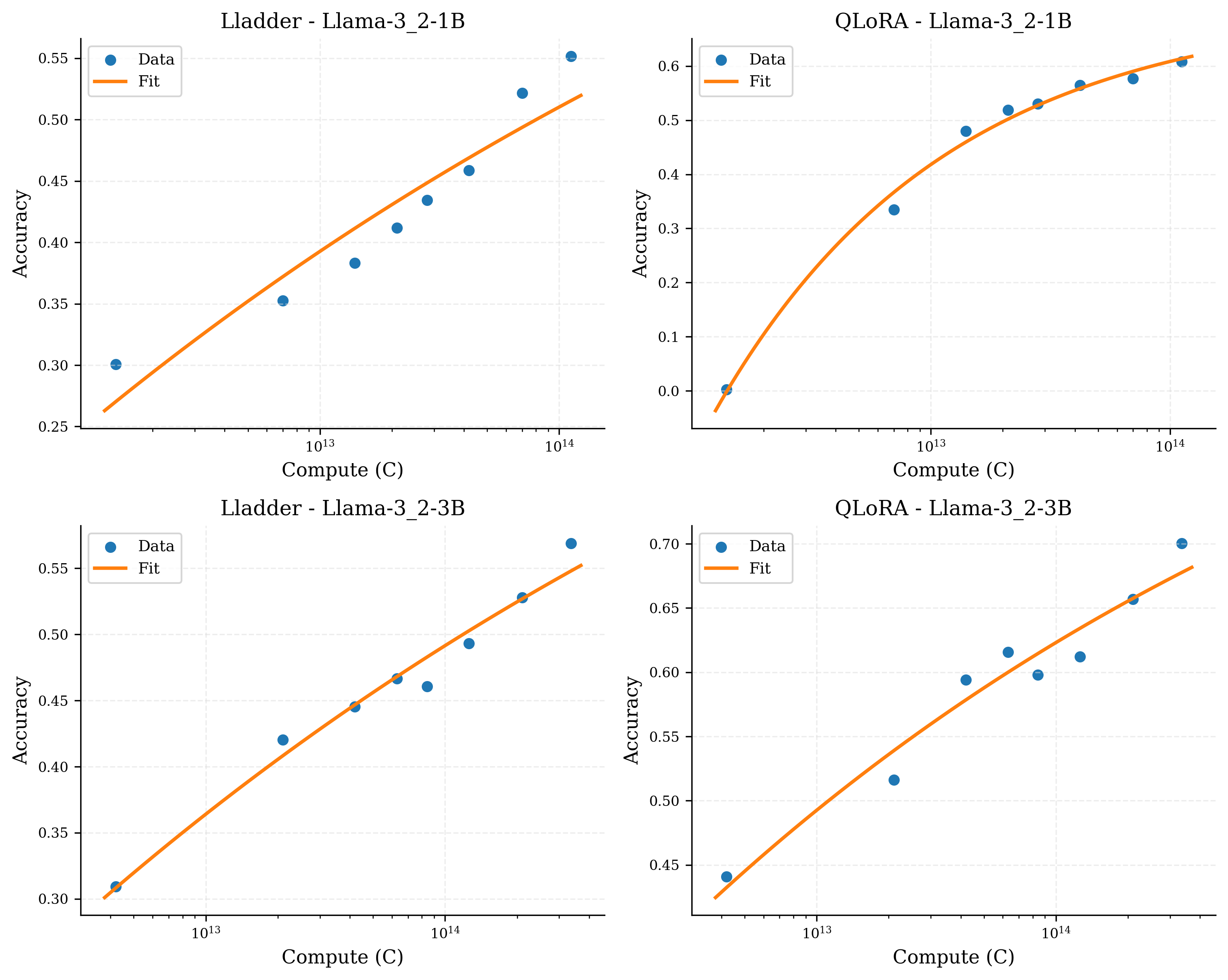}
\caption{Downstream scaling law for Llama series models comparing Ladder and QLoRA methods.}
\label{fig:llama_scalacc}
\end{figure}

Given the sparse sampling of compute, especially for QLoRA, the bootstrap intervals are broad; accordingly, Ladder and QLoRA exhibit similar scaling behavior within statistical uncertainty. All the fitted plots are shown in Figures~\ref{fig:qwen_scalacc},~\ref{fig:opt_scalacc}, and~\ref{fig:llama_scalacc}.

\subsection{Details on Scaling Laws}
\label{sec:isoflops}
\subsubsection{Intuitive Explanation of IsoFLOP}
Because our focus is performance as a function of compute, we follow the first scaling relation of \citet{kaplan2020scalinglawsneurallanguage}. For text generation, we measure test loss $Y(C)$ at increasing fixed compute budgets $C$ under two idealized assumptions: effectively unbounded data $D$ and unbounded model size $N$.

\begin{itemize}
    \item \textbf{Infinite data.} Each training is run for a single epoch, over as many data samples as possible until the target compute budget $C$ is exhausted.
    \item \textbf{Infinite model size.} For each $C$, we search over model sizes $N$ and take $N^*$ that minimizes $Y(C)$. In practice, larger $N$ increases per-sample cost, so at fixed $C$ a larger $N$ implies fewer training samples $D$.
\end{itemize}


Varying $N$ at fixed $C$ yields an \textbf{isoFLOP} test-loss curve with a characteristic U-shape: small $N$ underfits (insufficient capacity), while very large $N$ overfits the compute budget (too little data), both producing higher loss. The compute-optimal frontier is then obtained by taking, for each $C$, the minimum test loss along its isoFLOP curve.

\subsubsection{Practical Implementation of IsoFLOP for xLadder}

The “infinite model size” assumption is ill-suited to parameter-efficient methods such as LST: on isoFLOP curves, the compute-optimal $N^*$ typically lies far beyond reasonable ladder sizes. In practice, the best points would require very large adapter networks, undermining memory budgets and the premise of parameter efficiency.

To empirically trace an isoFLOP curve, we increase the effective parameter count $N$ by (i) adding non-connected extension layers and (ii) widening their hidden size.

\noindent\textbf{Setup.}
We use a Qwen2.5-14B backbone (48 layers, hidden size 5120). Each ladder layer mirrors a Qwen2.5 block with a reduced hidden size of 320. The Ladder has a depth of 5, connected to backbone layers $\{0, 11, 23, 35, 47\}$. We then insert $l$ non-connected extension layers immediately before the last connected ladder layer (i.e., 47), varying $l \in \{0,10,20,30,40,50,60,70,80,90,100,200\}$.

We fine-tune on a subset of the French Claire Dialogue corpus\footnote{\url{https://huggingface.co/datasets/OpenLLM-France/Claire-Dialogue-French-0.1}} to adapt the backbone to French dialogue. We select the first 10k utterances shorter than $4\times 2048$ characters and evaluate test loss on the processed test split, yielding 10 points along the scaling curve.

\begin{center}
\begin{figure}[htbp]
    \includegraphics[width=0.9\linewidth]{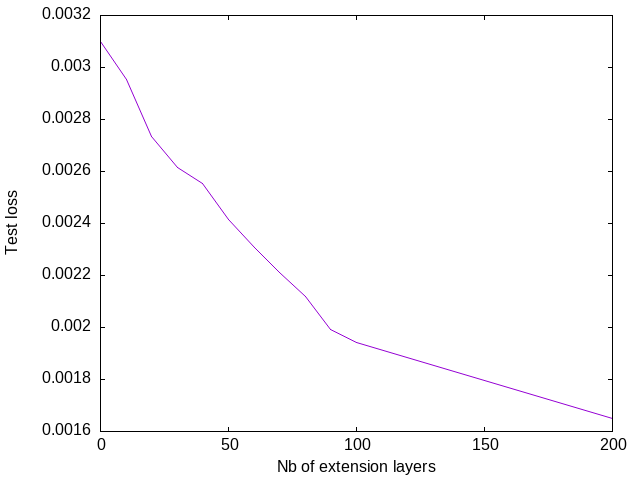}
    \caption[]{IsoFLOP of the xLadder.}
    \label{fig:isoFLOP}
\end{figure}
\end{center}

\noindent\textbf{Results.}
A representative isoFLOP curve is shown in Fig.~\ref{fig:isoFLOP}. VRAM usage grows with $l$, and we stop when out-of-memory (OOM) occurs on a 32 GB GPU.

As shown in Figure~\ref{fig:isoFLOP}, memory constraints within a parameter-efficient setting restrict us to the left side of the isoFLOP curve. Consequently, we select the largest feasible $N$, which is not compute-optimal but is the closest attainable point to the true optimum $N^*$.

Following prior work, we assume that post-Ladder performance follows a compute scaling law, analogous to full and other PEFT fine-tuning regimes. We cannot fully validate this assumption here: doing so would require broader sweeps over \textit{models}, \textit{datasets}, and \textit{scales}, plus out-of-fit extrapolation tests to assess generalization of the fitted law. Given limited compute, we adopt the standard scaling formulation and fit it for Ladder fine-tuning. This is a reasonable working assumption, supported by extensive evidence of scaling behavior in related training settings.

\section{Memory requirements computation}
\label{sec:appendix-vram}
We formalize peak per-GPU memory during fine-tuning under two regimes:
(i) QLoRA, and (ii) Ladder (Side) Tuning. We decompose peak memory into
frozen model parameters, trainable parameters, optimizer state, gradients, and
activations:
\begin{equation}
\label{eq:peak_total}
    M_{\mathrm{peak}} =
    M_{\mathrm{model}} +
    M_{\mathrm{optim}} +
    M_{\mathrm{grads}} +
    M_{\mathrm{acts}}
\end{equation}
We rely on the formulas given in Eleuther blog post 
\footnote{see \url{https://blog.eleuther.ai/transformer-math/\#training}} 
and section 4.1 of NVIDIA activations paper 
\cite{korthikanti2022reducingactivationrecomputationlarge} to compute the memory 
requirements of a transformer model during training assuming a vanilla attention 
mechanism and without a gated linear unit (GLU) or other non-linearities.

As stated by the Falcon Team \cite{almazrouei2023falconseriesopenlanguage}, for GLU the memory requirements are roughly x1.5 times larger, as it doubles the number of parameters in the MLP.

We assume a training in bf16, a 8-bit AdamW optimizer, no gradient checkpointing.

\noindent\textbf{QLoRA Memory Requirements.}
The main frozen parameters are quantized in 4 bits (i.e., int4), while the additional trained parameters are stored in 2 bytes (i.e., bf16). The memory required for the parameters is:
\begin{align*}
    M_{\text{model}} &= M_{\text{frozen}} + M_{\text{trainable}} \\
    M_{\text{model}} &= \frac{N}{2} + 2\,n
\end{align*}
with $N$ number of main (frozen) parameters and $n$ number of additional trainable parameters.

Classic AdamW stores 12 bytes par trainable parameter corresponding to 4 bytes for parameters, 4 bytes for the momentum and 4 bytes for the variance. While, AdamW in 8-bit stores 6 bytes per trainable parameter, so the memory required for the optimizer is:
\begin{equation*}
    M_{\text{optim}} = 6\,n
\end{equation*}

Gradients are typically stored on the full backpropagation path and requires 2 bytes (i.e., bf16) per parameter:
\begin{equation*}
    M_{\text{grads}} = 2\,n
\end{equation*}

To avoid incurring extra costs during backpropagation, activations of all parameters are typically computed, during the forward pass and kept in memory to later compute the gradients; this requires:
\begin{align*}
    M_{\text{acts}} &= M_{\text{acts,baseline}} + M_{\text{acts,qlora}} \\
    M_{\text{acts,baseline}} &= s\,b\,h\,L\left(34 + 5\frac {as}{h}\right)
\end{align*}

For QLoRA activations, each adapted linear adds (conceptually) the intermediate $Z = X A \in \mathbb{R}^{(b s)\times r}$, with $n=2hrmL$ trainable parameters then:
\begin{align*}
    A_{\text{QLoRA}} &= s\,b\,r\,m\,L \\
    M_{\text{acts,QLoRA}} &= 2\,A_{\text{QLoRA}} = \frac{n}{h} s\,b
\end{align*}
with $s=$ sequence length, $b=$ batch size (or micro batch size), $h=$ hidden size, $a=$ number of attention heads, $r=$ rank used in QLoRA.


Following \eqref{eq:peak_total}, the total memory required by QLoRA is:
\begin{equation}
\resizebox{\columnwidth}{!}{$
\label{eq:peak_qlora}
    M_{\text{peak,QLoRA}} = \frac N 2 + 10n + sbhL\left(34+ 5\frac {as}{h}\right) + \frac{sbn}{h}
$}
\end{equation} 

\noindent\textbf{Ladder Memory Requirements.}
We assume the same number of additional parameters $n$ in QLoRA and Ladder.
The memory for parameters and optimizer is the same.

Because there is no backpropagation in the main LLM, then the gradients memory is only:
\begin{equation*}
    M_{\text{grads, ladder}} = 2\,n 
\end{equation*}

But the main difference comes from the activations: they do not need to be stored at all in the main LLM, only in the ladder. Therefore, for the same low-rank dimension $r$, $l$ layers and $a_{lad}$ attention head in the ladder, we have:
\begin{align*}
    M_{\text{acts}} &= M_{\text{acts,ladder}} \\
    M_{\text{acts,ladder}} &= s\,b\,r\,l\left(34 + 5\frac {a_\text{lad}s}{r}\right)
\end{align*}

Following \eqref{eq:peak_total}, the total memory required by Ladder is:
\begin{equation}
\label{eq:peak_ladder}
    M_{\text{peak,ladder}} = \frac N 2 + 10n + sbrl\left(34+ 5\frac {as}{r}\right) 
\end{equation}  

Concretely, let's consider LLama3.1-70B: $N=70.10^9$, $h=8192$, $L=80$, $a=64$.
Its context length is $s=128,000$ tokens, and we assume $b=1$.
Let's assume adding $n=10^8$ parameters, i.e., 0.1\% of the parameters, with $r=8$, $a_{lad}=1$ and $l=L$ layers in the ladder.
Then, we have:

From \eqref{eq:peak_qlora}, we can compute the memory requirements for QLoRA:
\begin{align*}
    M_{\text{peak,QLoRA}} &= \frac{N}{2} + 10n  + \\
    &\quad sbhL\left(34+ 5\frac {as}{h}\right) + sb\left(\frac n {h}\right) \\
    &= 26.2\times 10^3s^2 + \\
    &\quad 22 \times 10^6 s + 37.75 \times 10^9
\end{align*}

From \eqref{eq:peak_ladder}, we can compute the memory requirements for Ladder:
\begin{align*}
    M_{\text{peak,ladder}} &= \frac{N}{2} + 10n + sbrl\left(34+ 5\frac {a_\text{lad}s}{r}\right) \\
    &=  400s^2 + 21.76 \times 10^3 s + 36\times 10^9 \\
\end{align*}

\noindent\textbf{Memory Requirements With Gradient Checkpointing.}

\begin{figure}[htbp]
    \centering
    \includegraphics[width=\columnwidth]{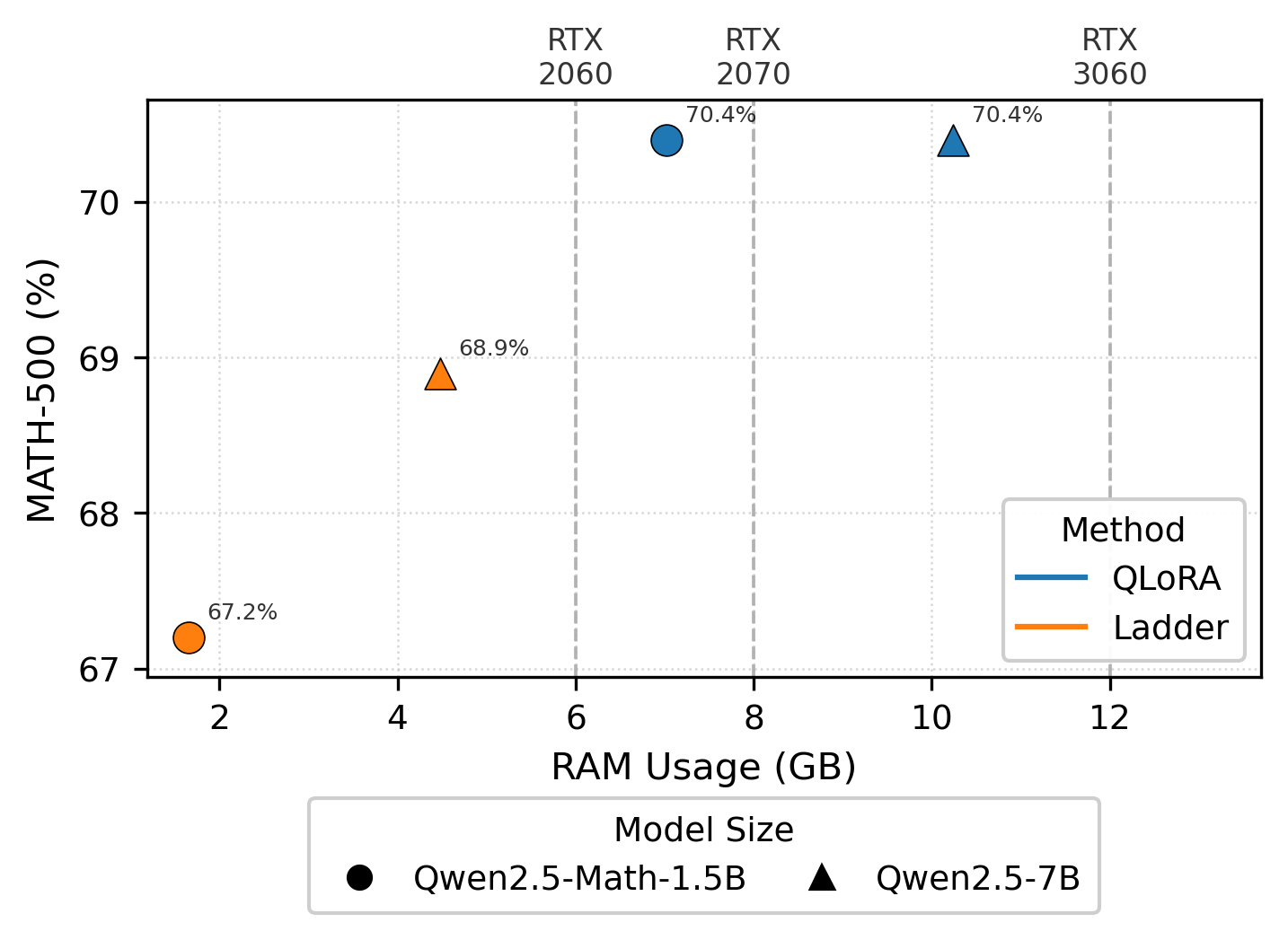}
    \caption{Accuracy on MATH-500 with gradient checkpointing enabled and a batch size of 16, with a 2k-token context length. RAM usage is computed for vanilla transformers, vertical lines represent which fine-tuning method can run on which consummer GPUs.}
    \label{fig:gradcheck}
\end{figure}

Gradient checkpointing, also known as activation checkpointing, trade memory for compute by freeing, during forward, all parameters' activations except at the output of every layer.
Then, during backward within each layer, this layer's activations are recomputed.

The above equations are thus simplified; for QLoRA, memory requirements are:
\begin{itemize}
    \item Activations: $2sbhL + sb(\frac n h)$
    \item Gradients: $2n$
    \item Parameters: $N/2 + 2n$
    \item Optimizers: $6n$
\end{itemize}

For (x)Ladder, memory requirements are:
\begin{itemize}
    \item Activations: $2sbrl$
    \item Gradients: $2n$
    \item Parameters: $N/2 + 2n$
    \item Optimizers: $6n$
\end{itemize}

Gradient checkpointing trades GPU memory for compute. For completeness, we report xLadder under checkpointing in Figure~\ref{fig:gradcheck}.
 
 
\noindent\textbf{Other Memory Optimizations.}
As explained in the Limitations section \ref{sec:limitations}, many various memory optimizations exist (e.g., FlashAttention, GQA, PagedAttention), they are complementary to Ladder architecture and can be used jointly with it.

A pipeline-style strategy is particularly well suited to fine-tuning with Ladder: stream one frozen backbone layer at a time over the corpus, write intermediate activations to disk, then train the (small) Ladder alone on these cached activations. This collapses GPU RAM to roughly a single-layer footprint at the cost of I/O and storage. In our setting, about 200 GB of fast NVMe sufficed for 30k sequences, enabling training on a very low-end GPU within a few days.

We limit ourselves in this work to vanilla attention and gradient checkpointing, leaving analysis of all other optimizations for future work.

\section{Implementation Math Reasoning hyperparameters}
\label{sec:appendix-implementation-math}
The training is performed for 3 epochs, and the model is evaluated on the validation set after each epoch. The sequence length is set to the maximum length of the dataset, which is 2122 tokens.
For Ladder fine-tuning, the batch size is set to 8 (resp. 4) for 1.5B model (resp. 7B model), and the learning rate is set to 2E-4. For QLoRA fine-tuning, the batch size is set to 4 (resp. 2) for 1.5B model (resp. 7B model), and the learning rate is set to 2E-5. The Ladder architecture used in this section is a fully connected ladder on each layer of the backbone model uniformly initialized without any gated sum, but only unweighted gated sum.
The backbone LLM is quantized in 8-bit precision. 
The model is trained on a single NVIDIA A100 GPU with 80GB of memory.

\section{Experiment Prompts}
\subsection{Math Reasoning Prompt}
\label{sec:appendix-math-prompt}
Only Qwen2.5 series models are used for math reasoning tasks. The prompt used is the same as the one used by \citet{yang2024qwen25mathtechnicalreportmathematical}.

\begin{promptbox}
    <|im_start|>system
    Please reason step by step, and put your 
    final answer within \boxed{}.<|im_end|>
    <|im_start|>user
    {problem}<|im_end|>
    <|im_start|>assistant
    {answer}<|endoftext|><|im_end|>
\end{promptbox}


\begin{promptbox}[title=Example]
    <|im_start|>system
    Please reason step by step, and put your 
    final answer within \boxed{}.<|im_end|>
    <|im_start|>user
    Given a rational number, write it as a 
    fraction in lowest terms and calculate 
    the product of the resulting numerator 
    and denominator. For how many rational 
    numbers between 0 and 1 will $20_{}^{}!$ 
    be the resulting product?<|im_end|>
    <|im_start|>assistant
    Given a rational number between 0 and 1, 
    we need to write it as a fraction in 
    lowest terms \(\frac{a}{b}\) and 
    calculate the product of the resulting 
    numerator and denominator. We need to 
    determine how many such rational numbers 
    exist where the product \(a \times b = 20!\).
    1. **Prime Factorization of 20!**:
        - The prime factors of 20! are 2, 3, 5, 7, 11, 
        13, 17, and 19. These are 8 distinct primes.
    
    2. **Coprime Pairs**:
        - For \(a \times b = 20!\) and \(\gcd(a, b) = 1\), 
        each prime factor of 20! must be assigned entirely
        to either \(a\) or \(b\). This results in 
        \(2^8 = 256\) coprime pairs \((a, b)\).
    
    3. **Counting Pairs \(a < b\)**:
        - Since \(a\) and \(b\) are coprime and 
        \(a \times b = 20!\), each pair \((a, b)\) and 
        \((b,  a)\) are distinct. To find pairs where 
        \(a < b\), we divide the total number of pairs 
        by 2, excluding the case \(a = b\) (which is 
        impossible since 20! is not a square).
    
    4. **Result**:
        - The number of coprime pairs \((a, b)\) with 
        \(a < b\) is \(\frac{256}{2} = 128\).
    
    Thus, the number of rational numbers between 0 and 1 
    for which the product of the numerator and 
    denominator in lowest terms is \(20!\) is 
    \(\boxed{128}\).<|endoftext|><|im_end|>
\end{promptbox}



\subsection{GLUE CoLA Prompt}

For the GLUE CoLA task, the CoLA sentence is directly given as a prompt to the LLM
complemented with a trained Boolean Classification head.

\begin{promptbox}[title=Example]
    User: Our friends won't buy this 
    analysis, let alone the next one we propose.
    Assistant: 1 (acceptable)
\end{promptbox}

\subsection{MATH500 LLM Critic Dataset}
\label{sec:appendix-llm_critic}
The MATH500 dataset is used in Section~\ref{sec:critic} to generate a new fine-tuning dataset
for the \textbf{LLM Critic} task. This task involves evaluating a given MATH-500
question, a model's CoT and its answer to determine if
the answer is correct or wrong.

The dataset is formatted as MCQ with two possible answers: 
the fine-tuned models has to generate a single token A for "Correct" or B for "Wrong".
This is enforced by applying the softmax only to both logits for "A" and "B".
The data is generated using Q4\_K GGUF versions of the following models via llama.cpp~\footnote{\url{https://github.com/ggml-org/llama.cpp}}:
\begin{itemize}
\item phi-4-Q4\_k\_gguf
\item Meta-Llama-3.1-8B-Instruct-Q4\_k\_GGUF
\item Qwen3-32B-Q4\_k\_GGUF
\item Qwen3-14B-Q4\_k\_GGUF
\item deepseek-math-7b-instruct-Q4\_k\_GGUF
\item Qwen2.5-32B-Instruct-Q4\_k\_GGUF
\item Qwen2.5-Math-7B-Instruct-Q4\_k\_GGUF
\end{itemize}

While the Q4\_K quantization may reduce the accuracy of model
reasoning, it enables creating the dataset at a reasonable computation cost.
The final dataset is balanced in terms of correct and incorrect answers, which are all contained within a `\textbackslash boxed\{\}` field.
All answers exhibiting severe repetition patterns are manually eliminated.


\begin{promptbox}[title=System Prompt all models except Qwen3]
    You are a highly intelligent and logical 
    mathematical assistant. Please reason step by step to 
    solve the problem. After your complete reasoning, put 
    your final numerical answer within \boxed{}. If the 
    answer is an expression, put the expression within 
    \boxed{}.
\end{promptbox}

\begin{promptbox}[title=Qwen3 System Prompt]
    You are a highly logical and accurate mathematical 
    reasoning engine. /think Please provide a detailed 
    step-by-step solution. Ensure your final answer is 
    enclosed within \boxed{}.
\end{promptbox}

\begin{promptbox}[title=User Prompt for MATH-500 Dataset Generation]
    Question: {problem}

    Please reason step by step to solve the problem.
    After your reasoning, put your final answer in 
    the format \boxed{your_answer}.
\end{promptbox}

\begin{promptbox}[title=Critic Task Prompt]
    User: {question}
    Assistant1: {generated model answer}

    User LLM Critic: Is the tentative answer 
    correct or wrong?

    Choices:
    A: Correct
    B: Wrong

    Assistant: {label}
    
\end{promptbox}

\begin{promptbox}[title=Example of Generation Prompt]
    You are a highly intelligent and logical mathematical 
    assistant. Please reason step by step to solve the 
    problem. After your complete reasoning, put your final 
    numerical answer within \boxed{}. If the answer is an 
    expression, put the expression within \boxed{}.

    Question: Convert the point $(0,3)$ in rectangular 
    coordinates to polar coordinates. Enter your answer 
    in the form $(r,\theta),$ where $r > 0$ and $0 \le 
    \theta < 2 \pi.$

    Please reason step by step to solve the problem. 
    After your reasoning, put your final answer in the 
    format \boxed{your_answer}.
\end{promptbox}

\begin{promptbox}[title=Example of LLM Critic]
    Two-thirds of the students at Baker Middle School 
    take music. There are 834 students who take music. 
    How many students are there at Baker Middle School?

    Here is a tentative answer to the math problem:

    Let's break down the problem step by step:

    1. We know that two-thirds of the students take music, 
    and there are 834 students who take music.
    2. Let's represent the total number of students at 
    Baker Middle School as x.
    3. Since two-thirds of the students take music, we 
    can set up the equation: (2/3)x = 834
    4. To solve for x, we can multiply both sides of the 
    equation by 3/2 to get rid of the fraction: $x = 834 
    \times (3/2)$
    5. Simplifying the right-hand side of the equation, 
    we get: $x = 834 \times 3/2$
    6. Multiplying 834 by 3 gives us: x = 2502 / 2
    7. Dividing 2502 by 2 gives us: x = 1251

    Therefore, there are 1251 students at Baker Middle 
    School.

    \boxed{1251}


    Question:


    Is the tentative answer correct or wrong?


    Choices:


    A: Correct

    B: Wrong


    Answer: A
\end{promptbox}

\end{document}